\pdfoutput=1

\documentclass[11pt]{article}

\usepackage[final]{acl}

\usepackage{times}
\usepackage{latexsym}
\usepackage{listings}

\usepackage[T1]{fontenc}

\usepackage[utf8]{inputenc}

\usepackage{microtype}

\usepackage{inconsolata}

\usepackage{graphicx}

\usepackage[utf8]{inputenc} 
\usepackage[T1]{fontenc}    
\usepackage{hyperref}       
\usepackage{url}            
\usepackage{booktabs}       
\usepackage{amsfonts}       
\usepackage{nicefrac}       
\usepackage{microtype}      
\usepackage{xcolor}         
\usepackage{multirow}

\usepackage{graphicx}
\usepackage{subcaption}
\usepackage{amsmath}
\usepackage{blindtext}
\usepackage{bbold}

\usepackage{tikz}
\usetikzlibrary{patterns}

\usepackage{paralist}

\newcommand{\emphtit}[1]{\textcolor{blue}{\textsc{#1}}}

\usepackage{algorithm}
\usepackage{algpseudocode}

\newcommand{\method}{\textsc{T-Free}}
\title{T-FREE: Subword \emphtit{T}okenizer-\emphtit{F}ree Generative LLMs via\\ 
Sparse \emphtit{R}epresentations for Memory-\emphtit{E}fficient \emphtit{E}mbeddings}

 \author{Björn Deiseroth$^{1,2,3}$  \quad
  Manuel Brack$^{2,4}$ \quad
  Patrick Schramowski$^{2,3,4}$ \quad
 \\
  \textbf{Kristian Kersting}$^{2,3,4}$\quad
  \textbf{Samuel Weinbach}$^{1}$ 
         \\
         $^1$ Aleph Alpha @ IPAI \hspace{.35cm} $^2$ Technical University Darmstadt\\
         $^3$ Hessian Center for Artificial Intelligence (hessian.AI) \\
          $^4$ German Research Center for Artificial Intelligence (DFKI) \\
         }

\begin{document}
\maketitle

\begin{abstract}
Tokenizers are crucial for encoding information in Large Language Models, but their development has recently stagnated, and they contain inherent weaknesses. Major limitations include computational overhead, ineffective vocabulary use, and unnecessarily large embedding and head layers. Additionally, their performance is biased towards a reference corpus, leading to reduced effectiveness for underrepresented languages.
%
To remedy these issues, we propose \method\, which directly embeds words through sparse activation patterns over character triplets, and does not require a reference corpus. 
\method\ inherently exploits morphological similarities and allows for strong compression of embedding layers. 
In our exhaustive experimental evaluation, we achieve competitive downstream performance with a parameter reduction 
of more than 85\% on these layers.
Further, \method\ shows significant improvements in \mbox{cross-lingual transfer learning.}

\end{abstract}


\begin{figure*}
    \centering
    \begin{subfigure}[b]{.46\textwidth}
         \centering
\includegraphics[width=\linewidth,clip,trim={0 0 40cm 0}]{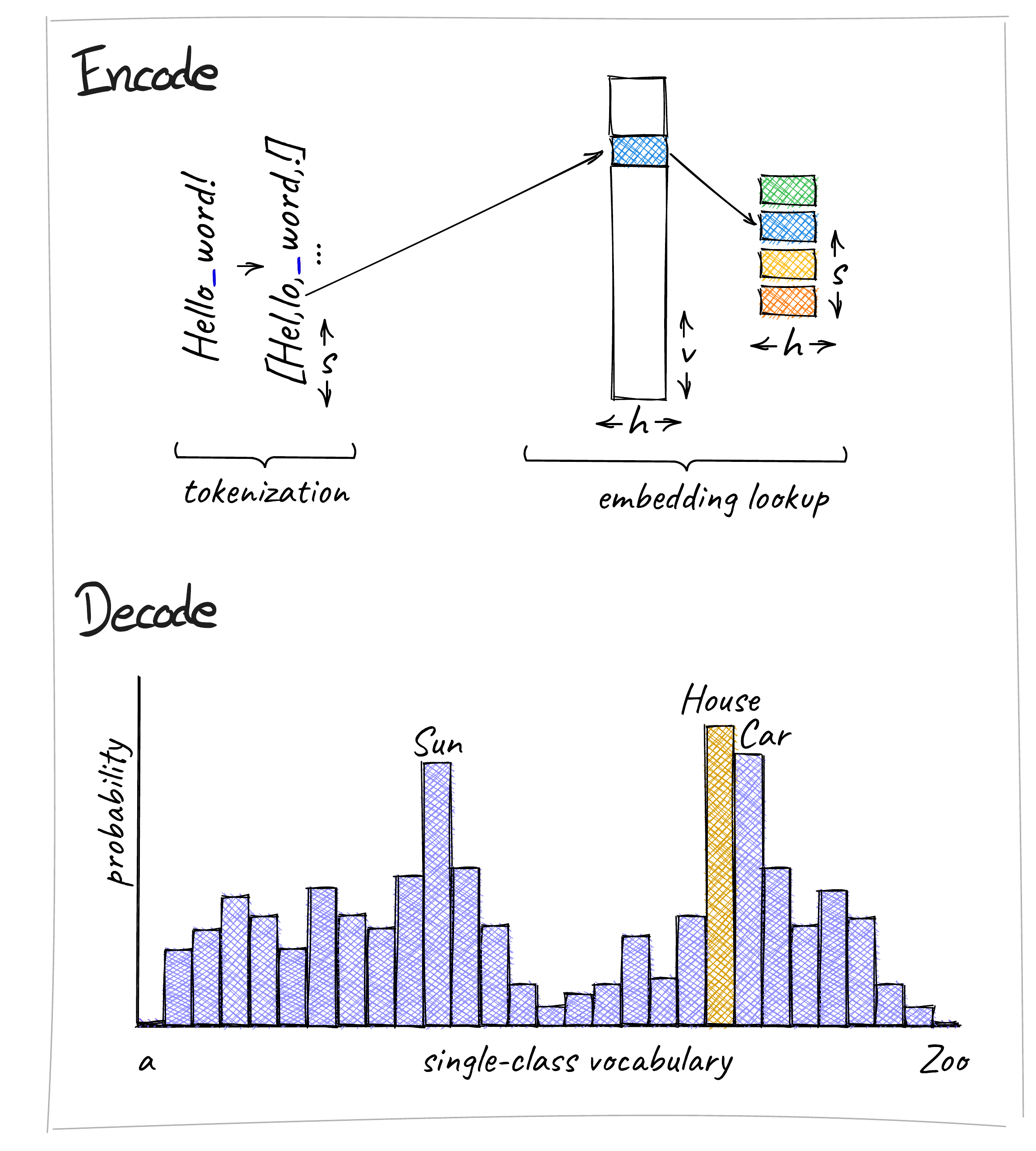}
\caption{Classic Tokenizer.}
         \label{fig:tokenization_endecode}
\end{subfigure}
    \begin{subfigure}[b]{.46\textwidth}
         \centering
\includegraphics[width=\linewidth,clip,trim={40.5cm 0 0 0}]{images/figure1_new.pdf}
         \caption{\method.}
         \label{fig:mesa_endecode}
\end{subfigure}
    \caption{Method comparison of  classic Tokenization (left) and \method\ (right) for text encoding (top) and decoding (bottom). Classic subword tokenizers learn a single-label vocabulary, i.e. a token is bijectively mapped into a single entry of the vocabulary. Instead, \method\ uses a bijective multi-label mapping over multiple activations of hashed character trigrams. As \method~explicitly models morphological similarities, it enables compression of the embedding layer.}
    \label{fig:endecode}
    \vskip -.5em
\end{figure*}

\section{From Text Representations For Machine Learning}
\label{sec:intro}

Large language models (LLMs) have shown remarkable abilities in processing natural language and various data types. The tokenizer, an essential part of any language-based LLM, splits input text into subwords and converts textual data into integer representation.
It is built by populating a fixed-size vocabulary based on statistical frequencies in a reference corpus \cite{sennrich2016bpe, kudo2018sentencepiece}.
With the LLM's trained embedding layers, these integers are converted into floating-point representations \cite{mikolov2013distributed, press2017using, vaswani2017attention}. These components significantly shape the training objectives and influence  what an LLM can process, interpret, and generate. Despite advances, the basic principles of tokenization and embeddings have remained largely unchanged in recent years.

{\let\thefootnote\relax\footnotetext{\url{https://github.com/Aleph-Alpha/trigrams}}}


Although this approach has served the LLM community well, 
and influential characters target to tokenize all kinds of data to ``lead a new industrial revolution''\footnote{\tiny{\url{https://x.com/tsarnick/status/1801884651986030820?s=12&t=5I__mymj5rXz7lxfplR8Gg}}},
it has significant inherent weaknesses. For one, subword tokenizers require dedicated training and, as such, additional computational resources. Design choices and errors at this stage can negatively impact the downstream model \cite{ali2024tokenizer}. 
Any tokenizer's vocabulary is heavily optimized for the reference corpus, leading to strong drops in performance for, e.g., underrepresented languages. 
We also show that the resulting vocabulary of subword tokenizers is poorly utilized, where up to 34\% of tokens are near duplicates with limited additional information. Despite that, the corresponding embeddings are trained independently. 
These issues have caused a significant expansion in the size of vocabularies and their corresponding embedding layers, with billions of parameters being allocated exclusively for text encoding and decoding.

To remedy these issues and challenge the traditional views, we propose a paradigm shift on how we embed and decode text for LLMs. We present tokenizer-free sparse representations for memory-efficient embeddings (\method) as an alternative to subword tokenizers. We directly embed each word in the input text with sparse activation patterns over hashed character triplets. Consequently, we eliminate the need for subword tokens, thus retaining near-optimal performance across languages. 
Additionally, \method~explicitly models character overlaps between morphologically similar words without the need to learn an embedding for each variant from scratch through a one-to-one bijection. The backbone of the language model will remain free of subword tokenization as we directly encode the textual representation. 
We argue that the converged embedding of such similar words should remain close and, thus, can be heavily compressed\footnote{The English language contains about $500k$ words, while ``native fluency'' is achieved at $10k$ words \cite{nation2006large}.}.
This exploitation of similarities allows \method~to reduce the size of the embedding layers by $87.5\%$\footnote{Compared to our $64k$ unigram baseline.} and the average encoding length of text by $56\%$\footnote{Compared to Mistral $32k$ avg. of EN, DE, RU, VI, AR.}. In addition to the inherent benefits of \method, the approach remains highly competitive on standard downstream model performance benchmarks. 
Finally, for transfer learning to an unseen language, the \method\ model quickly improves performance, while the tokenizer baseline shows only minor adaptation.

Our contributions can be summarized as follows: 
\begin{itemize} 
    \item We systematically demonstrate the inherent weaknesses of common tokenization and embedding approaches. 
    \item We propose \method, a powerful and efficient alternative for tokenizer-free LLMs.
    \item We exhaustively evaluate hyperparameters of \method~on established benchmarks by training 1B LLMs from scratch. Our comparison against equally trained models with classic tokenization demonstrates competitive performance despite the significant reduction in compute resources and parameters.
    \item We demonstrate the capabilities of \method~for cross-lingual transfer on continual pre-training 
    on a 3B LLM. 
\end{itemize}

\section{Classic Tokenization Principles}
\label{sec:background}
Before we derive \method~in detail, let us first establish some basics of how LLMs traditionally encode and decode text. Most LLM operations
are performed in floating-point numbers through a series of matrix multiplications and non-linear activation functions.
Consequently, we require techniques that map discrete textual inputs into floating-point representations and inversely transform the predictions of the model back to text. 

Traditionally, the first step in this process is to split any textual input into small chunks referred to as \textit{tokens}. Generally, these tokens can take arbitrary formats, spanning numerous characters, a single or even multiple words, and may also contain special characters. The latter can be particularly useful to encode programming languages. 
A \textit{tokenizer} comprises the steps and rules that are necessary to dissect a text into a sequence of tokens. Importantly, the total number of tokens is restricted, and we refer to the set of \mbox{all unique tokens as the \textit{vocabulary}.} 

Each token in the vocabulary is assigned an integer \textit{token-id}, wherefore tokenizers produce a sequence of token-ids for any textual input. 
Next, a large matrix of dimensions \emph{vocab size $\times$ hidden size}, an LLM's \textit{embedding layer}, maps each token-id to an internal representation as a floating point vector (cf. Fig.~\ref{fig:tokenization_endecode}). 
%
To produce new text, generative models are trained auto-regressively. That is, they iteratively predict the next token, which is appended to the input text. 
Therefore, the training objective is formulated as a classification problem: a one-label prediction of the next token over the entire vocabulary. 
Consequently, the last layer of the model---the \textit{LM head}---is a projection into the size of the vocabulary and thus also of dimension \emph{vocab size $\times$ hidden size}. For decoding, we can, for example, always select the token with the highest assigned value, which is called \textit{greedy sampling}.
The output text is produced by looking up the corresponding text snippet of each predicted token-id in the vocabulary.

Generally, it is desirable to encode any text in as few tokens as possible to reduce computational cost. 
At the same time, different semantic concepts should be separated into distinct tokens to ensure good language comprehension. The combination of both objectives is usually best satisfied by encoding each word as one token. 

\subsection{Tokenizer Algorithms}

The vast majority of LLMs utilize a tokenizer built with one of two approaches. Both progressively build up tokenization rules and their vocabulary based on statistics in a reference corpus.

\textbf{Byte Pair Encoding (BPE).}
BPE \cite{sennrich2016bpe} starts with a set of all characters as individual tokens. Progressively, the most frequent token pairs occurring together in the training documents are merged.
The resulting new token and the merging rule are added, and the training is completed \mbox{when the desired number of tokens is reached.}

In order to encode text with the trained tokenizer, BPE splits the input into individual characters and applies the lowest-ranking merge rule until no more are applicable. This exhaustive search can become computationally intensive, especially for long input sequences and large vocabularies. 

\textbf{Unigram.}
%
Unigram~\cite{kudo2018sentencepiece} operates inversely to BPE. 
First, it splits the training corpus into a large set of reference words and their respective frequencies. The vocabulary is initially populated with all possible substrings of these words.
At each iteration, Unigram computes a loss of the current vocabulary with respect to the training corpus for all possible tokenizations. The least influential tokens are then removed until the desired vocabulary size is reached. 
%
For text encoding, the Viterbi algorithm is applied to determine the most preferred segmentation of a given word based on the ranked available tokens.

The text decoding in both cases maps directly back into the vocabulary list and the respective sub-words. 
To ensure that every word can be represented, a ``byte-fallback'' into unicode is often used for characters not present in the vocabulary.

\subsection{Facing the Flaws} \label{sec:flaws}


Common to both methods is a set of distinct flaws.

\textbf{Large Vocabularies F1)} Words that do not appear in the vocabulary are split into multiple tokens and, as such, require more compute during model inference and training.
To avoid out-of-vocabulary words and to achieve the best downstream representations on a diverse set of languages and tasks,
researchers tend to use ever larger vocabularies. Although some models still rely on a $32k$ vocabulary \cite{touvron2023llama2, jiang2023mistral}, more recent releases go up to $128k$ \cite{meta2024llama3} or even beyond $250k$ \cite{mesnard2024gemma, cohere2024commandr}. 
Large vocabularies, in turn, require large embedding and head layers. 
For example, Command-R \cite{cohere2024commandr} with a hidden dimension of $12,288$ and a vocabulary of $256,000$ tokens uses $6.3B$ parameters only for the embedding and head layer.
Naturally, a large number of parameters complicate model training and may require 
advanced sharding techniques such as ``model parallelism''.
Even the tokenization itself can become \mbox{(CPU-)} computationally intense for large documents and vocabularies.
Naturally, embedding matrices of this scale are generally not an option for smaller ``on-the-edge'' models. 
Nevertheless, they still occupy a large portion of parameters in smaller models, e.g. 40\% for Gemma-2B \cite{mesnard2024gemma}.

\textbf{Duplicate Tokens F2)} Furthermore, the allocated vocabulary is expected to be poorly utilized due to the statistically likely occurrence of near-duplicate tokens. 
Most prominently, a significant portion of tokens appears multiple times, only differing in capitalization or the existence of a leading whitespace (\emph{cf.} Sec~\ref{sec:exp_flaws}).
For example, to spell all 64 substrings and variations of the word ``\_words''\footnote{$\_$ represents a whitespace.}, we require a total of 37 unique tokens (\emph{cf.} App.~Tab.~\ref{tab:word_prefixes}). 
Since the corresponding embeddings of all tokens are independent and randomly initialized, the representation of each duplicate token needs to be learned from scratch without exploiting morphological synergies.
Further, large embedding layers are purely utilized since some tokens will rarely occur. The corresponding embedding weights of these tokens are thus seldom active while still requiring compute.

\textbf{Training data overfitting F3)} As discussed above, these tokenizers require dedicated training before the actual model training. In addition to the added computational overhead, the data selection and potential mistakes during tokenizer training have significant impact on the subsequent LLM \cite{ali2024tokenizer}. 
For natural language, for example, this paradigm may result in a vocabulary tailored to one language (usually English) and consequently drops in performance for others, especially those not explicitly included. 
The resulting LLM may still be somewhat adapted to other languages since many similar low-level structures \cite{nikolov2013exploiting}. However, its overall training and inference performance will not be as efficient as we demonstrate.


In contrast, \method~addresses all of these disadvantages. It is computationally efficient and performs good tokenization across languages without duplicates. It drastically reduces the parameters required for text encoding, exploiting word spelling similarities. Importantly, none of these improvements sacrifices downstream model performance.







\section{\method}
\label{sec:trigram}
A key motivation for \method~is the intuition that minor differences in spelling, like leading whitespaces or capitalization, do not hold enough entropy to justify entirely independent tokens. 
\method~directly encodes morphological similarities by representing each word as a multi-label encoding of its character triplets. This designed overlap between words allows us to significantly reduce the size of embedding layers.

We now derive \method's approach to text encoding and decoding and discuss implications on LLMs in general. We provide a visualization of each step in Fig.~\ref{fig:mesa_endecode} and pseudo-code in App.~\ref{app:mesa_algo}.

%



\subsection{Text Encoding}
\label{sec:encoding}
\textbf{Step 1: Word splitting.} First, we rigorously split the text by digits and non-alphanumeric characters. The resulting splits, therefore, contain entire words, digits, or special characters.
We consider each digit separately, as it is standard in SOTA LLMs (\emph{cf.} Tab.~\ref{tab:language_fertility}). Specifically, we include the 10 digits $0$ to $9$, and otherwise, we rely on attention to comprehend larger numbers or mixtures with characters.

By definition, we represent each word with a prefixed and suffixed whitespace. In particular, we assume that an entire word is encoded into a single embedding, and analogously, we predict an entire word at once.
Consequently, we no longer need to explicitly model whitespace as a character and eliminate near-duplicate tokens. 
Nonetheless, we add a dedicated ``whitespace'' and ``non-whitespace'' token to the tokenizer. These special tokens allow us to model cases where substrings should (not) be concatenated with whitespace, e.g., single digits of larger numbers. 
To reduce their need, we further add a rule-set that favors \mbox{(non-)}whitespace in front or after certain characters. Generally, we prefer to add no whitespace after a digit embedding and similarly no whitespace before punctuation.  
A detailed description of the rule set is found in App.~\ref{sec:whitespace}.

Considering the example in  Fig.~\ref{fig:mesa_endecode}, the input text ``Hello\_word!'' would be tokenized as [`Hello',`word',`!'].

\textbf{Step 2: Encoding.} Next, we define a robust hash function that uniformly encodes a token into $n$ descriptors, where $n$ usually equals the word-length\footnote{Only exceptions are unicode fallbacks.}.
Specifically, we apply convolutions of size three and byte-wise stride to each word. This operation yields a set of character triplets, which we refer to as ``trigrams''. 
Consequently, ``Hello'' is decomposed into \{\_He,Hel,ell,llo,lo\_\}.
Trigrams usually contain enough information about the relationship between letters to reassemble the word from the unordered set. 

Subsequently, we project each trigram descriptor into a sparse hidden representation vector of $m$ ``active entries'' on the embedding layer. 
Specifically, \method~calculates $m$ numerical hashes of each trigram, which can be considered as identifiers.
We map these into the LLMs embedding matrix by calculating each hash value \texttt{modulo} $v$ to identify the active indices. The selection of vocab size $v$ is further explained in Step 3.

Overall, we obtain $n\cdot m$ total activations for any single word.
To further exploit word similarities and bootstrap training, we calculate $k\in [0,m)$ out of these hash calculations with the lowercased trigram.
This mapping from trigram to hidden representation is static and can be precomputed\footnote{Note that there are only $256^3\approx 16.7M$ trigrams.}.

\textbf{Step 3: Aggregation.}
Similar to classic embedding approaches (\emph{cf.} Fig.~\ref{fig:tokenization_endecode}) \method~also utilizes an embedding matrix of dimension $v$ with hidden size $h$. However, we do not have a fixed vocabulary, whose size dictates $v$. Instead, we can independently choose $v$ as a hyperparamter with words and trigrams sharing individual entries to better encode similarities. 
Lastly, we sum all $n\cdot m$ embedding entries to produce the final one embedding corresponding to a word, such as ``Hello''.

Note again, that we utilize a significantly smaller number of embeddings than there are trigrams. While their hashes may naturally overlap, we ensure the uniqueness of the  entire patterns through the $m$ simultaneous hashes. As we ensure that trigram encodings do not collide, neither will the word encodings.

\subsection{Training Objective \& Text Decoding}

As \method's representation of a word is now a multitude of activations, we reflect this change in the LM head, as well (\emph{cf.} \textit{Decode} sections in Fig.~\ref{fig:endecode}, App. Alg.~\ref{alg:encode},\ref{alg:decode}).
In particular, we change the target loss function from classic single-label binary cross-entropy (BCE) to a multi-label (ML) BCE over all $n\cdot m$ activations of the next word targets: 
\begin{equation*}
\small\mathcal{L}^{ML}_{BCE} \small\textstyle = - \sum_{j=1}^{v} [y_{j} \log(\hat{y}_{j}) + (1-y_j)\log(1-\hat{y}_{j})],
\end{equation*}
for $\hat{y}$ being the LM's prediction and  $y$ the binary  target vocab labels with $\sum_{j=1}^{v} y_j  =  n\cdot m$.

\begin{figure}
\centering
\includegraphics[width=\linewidth,clip,trim={0 15cm 24cm 0}]{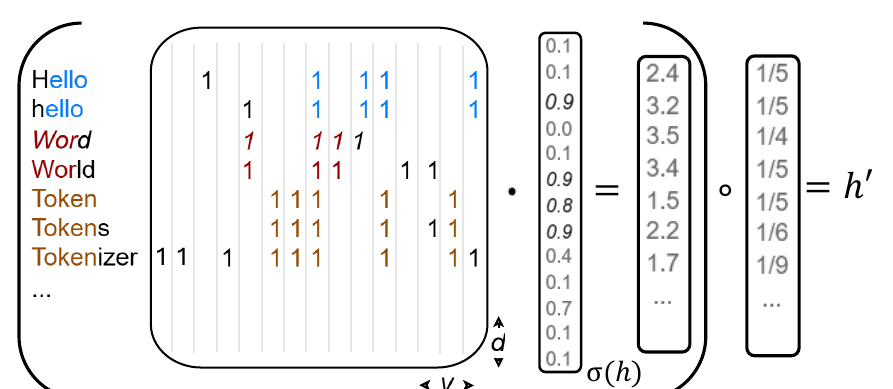}
\caption{Example of the next word prediction with \method. To the list of predictable words of dimension $d$ we generate once the corresponding patterns within the available vocabulary size $v$, as described in the encoding step $2$ of Sec.~\ref{sec:encoding}. Note how morphologically close words will generate overlapping patterns. The element-wise sigmoid values of the output of the last hidden layer, $\sigma(h)$, is multiplied with this pattern matrix using standard dot product. Finally, we use $h'$ for the sampling process, the average sigmoid value of a word. C.f. App.~\ref{app:mesa_algo} for further details.}
\label{fig:decode}
\end{figure}

Next token decoding is shown in Fig.~\ref{fig:decode}. We first assemble a dictionary of all possible next words and pre-compute their activation patterns.  
Importantly, only $n\cdot m$ out of $v$ entries will be non-zero for each word, and since we choose $m <\!< v$, the dictionary matrix can be encoded as a sparse matrix, thus improving runtime.
In addition, note the pattern similarity between similar words, as previously described.
The last hidden layers' output $h$ is sigmoided, and multiplied with the dictionary matrix.
Finally, we compute the average sigmoid value per dictionary entry, $h'$, to sample the next word, e.g. using standard argmax. Overall, for a dictionary with $512k$ entries, this procedure only marginally increases the decoding runtime due to the sparse property of the activation patterns. Further description, along with pseudocode, detailed depictions, and step-wise runtime analysis can be found in App.~\ref{alg:decode}. 

Note that the decode matrix is not required during training, and can dynamically be exchanged. We generate it by sampling the top-$500k$ occurring words in the training dataset, and dynamically adding the missing words of the prompt. 

\subsection{Distinctions of paradigm shift}
Notably, this paradigm shift to a multi-class vocabulary allows for more semantically robust decoding. With the classical approach, the distinctly noisy learning process can lead to unrelated concepts appearing among the top predictions (\emph{cf.} `\textit{House}' and `\textit{Car}' in Fig.~\ref{fig:tokenization_endecode}). This effect can have a significant impact on next token sampling and potentially devastative outcomes for model modifications such as compression~\cite{deiseroth2024divergent}. 
In contrast, the trigrammification and resulting embedding overlap of similar words with \method~
inherently favors similar words during decoding  (\emph{cf.} `\textit{ouse}' in Fig.~\ref{fig:mesa_endecode}).
Moreover, activations in the embedding and LM head are more uniformly distributed, leading to better parameter utilization, and more stable model behavior. 

The predictable words are still derived from a dictionary. However, this vocabulary list is exchangeable, and is not required during training. As such, depending on the use-case, it may be kept in reasonable sizes. Moreover a hierarchical decoding exploiting morphological structures can straightforward be implemented, e.g. first decoding lowercase words, and then uppercase variations (or similarly grouping by stems or endings).


Lastly, our design of a robust hash function on words adresses the afore mentioned flaws (Sec.~\ref{sec:flaws}) as the results of the next section demonstrate.

\section{Empirical Evaluations}
We continue with an empirical demonstration of the performance of \method, and how it remedies the flaws of standard tokenizers as outlined in Sec.~\ref{sec:flaws}.  
To thoroughly analyze the performance differences, we designed three consecutive experiments:
First, we perform hyperparameter ablations on a series of 1B parameter models, which achieve competitive scores on standard benchmarks with a reduced vocabulary, which in turn addresses \textbf{F1}.
Second, we analyze the duplicates in the tokenizers of recent LLMs with respect to \textbf{F2}. 
Notably, \method\ is by design free of duplicates. 
Lastly, we look at \textbf{F3} and evaluate the performance of various tokenizers across languages. Further, we trained 3B parameter models on English and continued training on German data to practically investigate language adaptability.  \method\ has better tokenization performance across languages and outperforms classic tokenizers on language transfer.

\subsection{Experimental Details}
First, let us clarify some details about our experimental setup. We provide more details for each section in the Appendix.

\textbf{Data and Code.}
We use the slimpajama dataset~\cite{cerebras2023slimpajama} as our English and Occiglot Fineweb v0.5 \cite{brack2024occiglot}
as our German data corpus.
Both datasets contain a diverse range of content and have been extensively filtered and deduplicated. 

As a baseline, we trained BPE and Unigram tokenizers of sizes 32$k$ and 64$k$ on a random 20GB slimpajama sample using Sentencepiece\footnote{\url{https://github.com/google/sentencepiece}}. More details are described in App.~\ref{app:sentencepiece}.

To ensure fair comparisons, we trained 1B and 3B models from scratch for the baselines and \method\ using our adjusted code base\footnote{\url{https://github.com/Aleph-Alpha/trigrams}}.

\begin{figure}
    \centering    
         \includegraphics[width=.9\linewidth,height=140px]{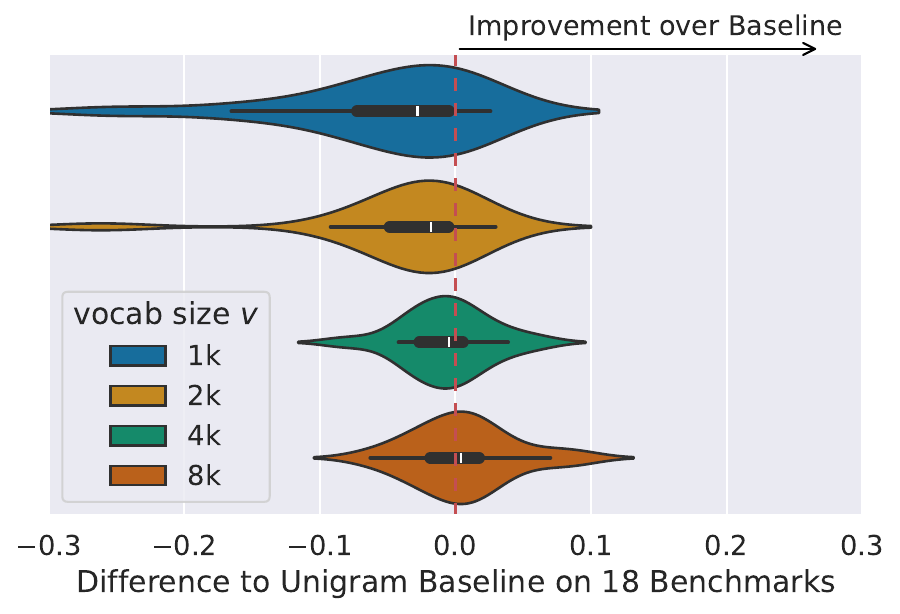}
         \caption{Hyperparameter search for Vocab Size of \method\ on a series of 1B ablations. We fixed number of activations $m=10$, and do not apply lowercase overlap ($k=0$). The boxplots show the differences of trained models to a $64k$ unigram baseline for 18 downstream benchmarks (0-shot). 
         \method\ outperforms in median the classical tokenizer architecture with a reduced vocab size of $8k$ entries ($12.5\%$).}
    \label{fig:hyperparameter_1b}
\end{figure}

\textbf{LLM Pre-Training.}
All models are transformer, decoder-only architectures similar to Llama-2. We solely change the tokenizer, embedding layer and LM head. Consequently, ablations with smaller sizes of $v$ result in a lower overall parameter count, heavily skewing the comparison in \textit{favor of the baseline}. 
For hyper-parameter ablations of \method, we train 1B models for $50k$ steps with $2k$ sequence length and $1k$ total batch size.
We then scale up the baseline and \method\ models to 3B parameters and train for $110k$ steps on slimpajama with $4k$ sequence length. For the multilingual learning experiment, we continue training this English 3B model at a lower learning rate for another $20k$ steps on German Occiglot data with a $20\%$ replay of English.  


\textbf{Evaluation.} 
We evaluate tokenizer performance in isolation using fertility measurements similar to \citet{rust2021howGood}. Fertility benchmarks the number of tokens required per word with $1.0$ thus being the optimal value.
Specifically, we 
compare different tokenizers across 5 diverse languages on the respective data from Wikipedia. 

Downstream benchmark comparisons are performed on 18 standardized benchmarks\footnote{\scriptsize\url{https://github.com/EleutherAI/lm-evaluation-harness}} in English that measure a wide variety of LLM capabilities, including general language modeling, question answering, and common sense reasoning. 
To evaluate german and english in comparison we use german translations of the Hellaswag, Truthfulqa and Arc-Challenge benchmarks\footnote{\url{https://github.com/bjoernpl/GermanBenchmark}}.

We built \method's prediction dictionary, from the top $80k$ words that occurring in slimpajama, and additional top $20k$ words from the German Occiglot data. 



\begin{table*}[h]
\small
    \centering
    \begin{tabular}{l | r r r r | r r r r r }
    \multirow{ 2}{*}{\textbf{Model/Tokenizer}} & \multicolumn{4}{ c |}{\textbf{Portion of duplicate tokens (\%) $\downarrow$}} & \multicolumn{5}{ c }{\textbf{Fertility across languages $\downarrow$}}\\
        &  Total & Cap. & Space & Digits & {EN} & {DE} & {RU} &{VI} &{AR}\\ \hline
      Unigram (64k)& 35.24 & 23.27 & 13.47  & 0.00 & 1.280 & 2.004 & 11.431& 5.060 & 9.455 \\
      BPE (64k)& 35.24 & 23.27 & 13.47  & 0.00 & 1.275 & 2.025 &  11.423 & 4.755 & 9.465 \\
      \hline
      Mistral (32k) & 31.47 & 19.10 & 16.45 & 0.00 &  1.397 & 1.931 & 2.560 & 3.346 & 4.722 \\
      Phi-2 (50k) & 23.23 & 12.91 & 16.89 & 3.32 & 1.265  & 2.266 & 6.729 & 4.339 & 5.225 \\
      Gemma (256k) & 34.68 & 20.27 & 20.50 & 0.04 & 1.176 & 1.447 &  1.903 & 1.726 & 1.793 \\
     
      Command-R (255k) & 15.31 & 15.31 & 14.00 & 0.00 & \textbf{1.152} & 1.411 &  1.590 & 1.597 & 1.578 \\
      \hline
      \textbf{\method} (Ours) & \textbf{0.00} & 0.00 & 0.00 & 0.00  &  {1.163} & \textbf{1.182} & \textbf{1.338} & \textbf{1.400} & \textbf{1.086} \\

    \end{tabular}
    \caption{Demonstration of inherent benefits of \method~over traditional tokenizers. The performance no longer degrades when confronted with languages beyond the one primarily trained on. Additionally, the vocabularies of classic tokenizers contain large portions of tokens only differing in their capitalization or leading whitespace. \method~does not construct such a vocabulary in the first place and thus utilizes embeddings more efficiently.}
    \label{tab:language_fertility}
    \vskip 1.em
\end{table*}

\subsection{\method\ performs at 8$k$ vocab size}
We present the results of our hyperparameter ablation study of \method\ for $1B$ models in Fig.~\ref{fig:hyperparameter_1b}. All scores are reported as differences to the Unigram $64k$ baseline and for fixed parameters $m=10$ and $k=0$. 
Generally, \method~remains highly competitive with the baseline as all versions outperform the Unigram model on some of the benchmarks. Further, we achieve the best results for a vocab size $v$ of $8k$ at which \method~outperforms the baseline on average. 
In contrast, a vocab size of $\leq 2k$ seems insufficient with devastating outliers.
We performed further ablations on parameters $m$ and $k$, which are outlined in App.~\ref{app:hyperparameter_ablation}. 

These results demonstrate that \method\ successfully addresses the flaw of large vocabularies and embedding layers (\emph{cf.} \textbf{F1} in Sec.~\ref{sec:flaws}).  
We are able to achieve competitive performance with only $12.5\%$\footnote{$8k$ instead of $64k$.} of the embedding parameters  using \method~instead of Unigram. 

Note, that we do not adjust any other model parameters when reducing vocab size. 
As such, the benchmark results compare a Unigram model with $1.07B$ parameter against a \method\ model with $0.84B$ parameters (for $v=8k$). 
 Consequently, we demonstrate that an LLM using \method~instead of Unigram performs better, despite having over 20\% fewer parameters.

\subsection{\method\ no duplicates by design}
\label{sec:exp_flaws}
Let us now look into (near) duplicate tokens  in commonly used tokenizers (\emph{cf.} \textbf{F2} in Sec.~\ref{sec:flaws}). In general, there are three types of overlaps in vocabularies: 1) The same token with and without capitalization, 2) with and without leading whitespace, and 3) dedicated tokens for multiple digits. 

In Tab.~\ref{tab:language_fertility}, we report the percentage of duplicate tokens for our baseline tokenizers and commonly used models. Overall, between 15\% and 35\% of the available vocabulary is spent on (near) duplicate information with limited differences in entropy. Generally, tokenizers contain the most duplicates for capitalization, slightly fewer for whitespaces, and only a few duplicate digits. The relative amount of overlap tends to decrease with larger vocabularies, although Gemma marks an inglorious exception.
In contrast, \method~is inherently designed to be free of duplicates.
We can even adjust the parameter $k$ to explicitly model the overlap of words to their lowercase representations. Consequently, all variants are inherently well represented in the emedding layer. 

\subsection{\method\ has lower fertility across, \mbox{and is more adaptive to new languages}}
\label{sec:main_results}

Finally, we investigate the versatility of tokenizers beyond their (main) language (\emph{cf.} \textbf{F3} in Sec.~\ref{sec:flaws}). We report the fertility of our baselines and other popular models in English, German, and three dissimilar languages that also contain significant character-level differences in Tab.~\ref{tab:language_fertility}. Common to all tokenizers is a significantly decreasing performance for non-English languages, especially for Russian and Vietnamese. Naturally, larger vocabulary sizes tend to have better multilingual coverage
, in particular to language groups close to English, but still suffer from significant performance drops. 
In comparison, the tokenization of \method, which is mainly based on whitespace splitting, provides comparably good performance across all $5$ languages\footnote{More detailed evaluations are found in App.~\ref{app:fertility}.}. The increases in fertility for Russian or Vietnamese remain small and there is no performance difference for German or Arabic. 
Note that these synergies were explicitly modeled, and no reference corpus is needed to train and bias the fertility of \method.
Consequently, \method~allows for easier and more efficient model adaptation to low-resource languages.

\begin{figure}
    \centering    
         \includegraphics[width=\linewidth]{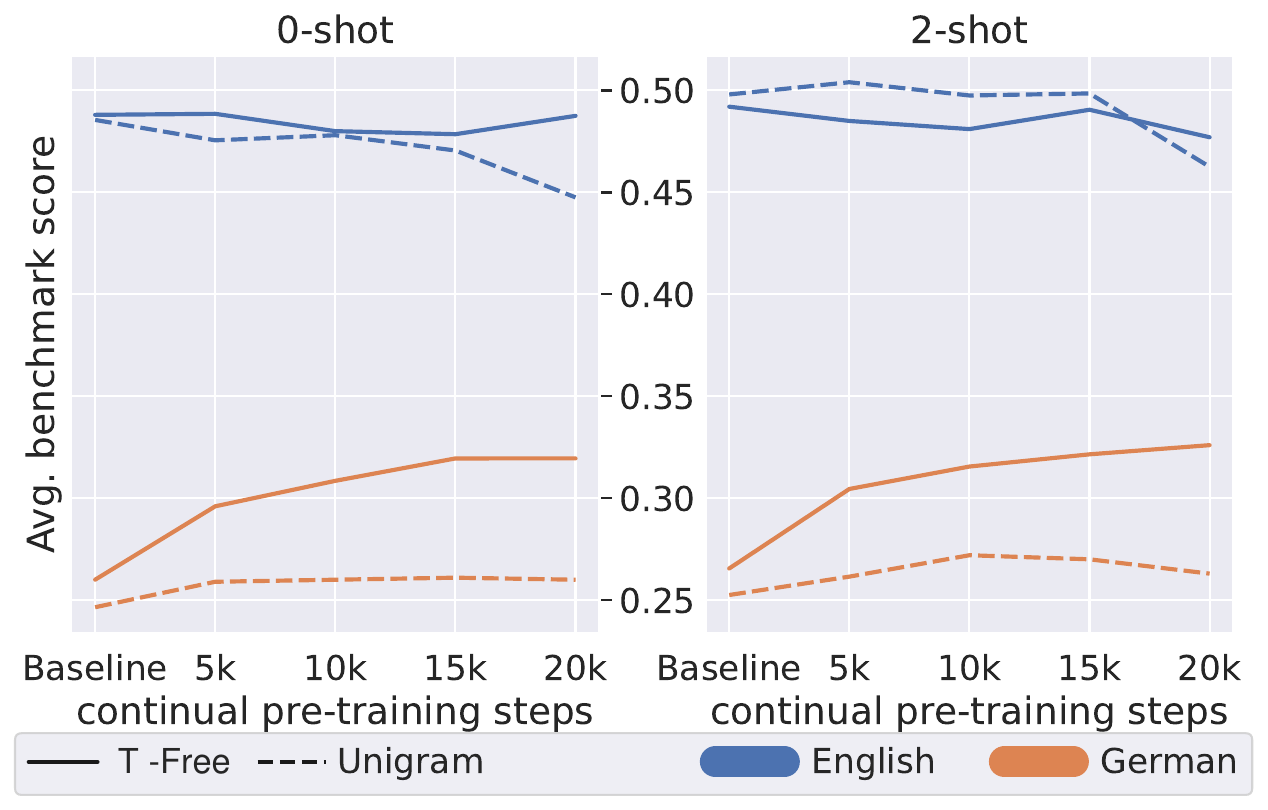}
         \caption{Continual pre-training performance. Trained are $3B$ models on English slimpajama data for $90k$ steps (``baseline''), and continued on German occiglot data for $20k$ steps. Plotted are the average scores of two benchmarks available in German and English: Hellaswag and Arc-Challenge. Notably, \method\ outperforms in German already with the baseline. Within $20k$ continued steps, \method\ improves by $5\%$ on average in 0 and 2-shot, while the classic tokenizer approach barely improves. Both models slightly drop performance in English, albeit the tokenizer version more drastically. Full evaluations are found in Appendix Tab.~\ref{tab:evalende},\ref{tab:evalen1},\ref{tab:evalen2}.}
    \label{fig:continual_3b_german}
\end{figure}

We now explicitly show the devastating consequences of biased tokenizers on the language transfer capabilities of LLMs. 
As discussed above, we first train $3B$ models for \method~and Unigram on English, and then transition to German.
Through more ablations, we fixed the activations to $m=7$ and the lowercase trigram overlap to $k=3$.
Fig.~\ref{fig:continual_3b_german} shows the performance average on the English and German versions of the standard benchmarks.
The baseline performance in German is already improved with \method, indicating that syntactic and semantic similarities between the languages are better captured in the learned representations.
Additionally,  \method\ almost achieves the English-level performance on German after $20k$ training steps. In contrast, the classical tokenizer variant improves only marginally with the same amount of training.

We, again, do not adjust any other model parameters when reducing the vocab size. 
As such, \method~uses 10\% fewer parameters than the baseline ($2.77B$ instead of $3.11B$) and still strongly outperforms the Unigram variant. More detailed evaluations are found in App.~\ref{app:benchmarks}.

\section{Discussion}
\label{sec:discussion}
Prior research has demonstrated that the mapping into a sparse hidden representation and the training of a dense aggregation layer as applied in \method, is a universal function approximator \cite{cmac}. These results provide further theoretical motivation for our approach.

 \method\ allows for significant compression of an LLMs' vocabulary by more than 85\% without performance degradation. 
 Notably, the affected embedding and head layers are by far the largest in LLMs in terms of parameter count. They are also the most influential to an LLM, as they dictate the mapping between text and numerical representations. For one, these massive improvements allow for better utilization of billions of parameters in large models.
 The compression of \method~in particular paves the way to building better low-resource models, by reducing model size and training cost and improving adaptability. 
 For example, in our experiments without pipe or model-parallelism, we were able to \emph{triple} the micro-batch size, yielding faster training iterations.
 

Furthermore, we observed more stable loss curves for \method, in particular for higher learning rates. These improvements may be attributed to the explicit modeling of similar words, the removal of duplicates, and the less volatile multi-label training target.
Further, the uniform hashing distributes gradients evenly amongst the available vocab size, in contrast to classical approaches.
We provide further details in App.~\ref{app:training_parameters},\ref{app:training_stability}.

The rules we use for obtaining word representations are universal and well-defined at pre-training time. They do not change over time, particularly neither when adding languages later on.
\method\ also lowers computational costs due to its low fertility and easy-to-process whitespace splitting. Consequently, pre-processing, training and inference of an LLM all require less compute. 

Lastly, \method\ allows to explicitly model and steer the decoding process at inference time, by altering the available dictionary. 
Consequently, hallucinations will likely be reduced due to fewer ``generic fall-back'' word splits. Moreover, one can dynamically add or remove words. 
It is worth pointing out that \method's compression benefits can also be combined with traditional tokenizers. Instead of the simple whitespace splitting one could keep traditional tokenization and trigramify ``classic tokens''. 

\section{Related Work}

Few alternatives to BPE and Unigram have been found in recent LLMs and research. 
\citet{charformer} propose a gradient-based trainable tokenization module in contrast the otherwise statistical based approach.

The naive approach of splitting the input text into bytes or characters maximizes fertility and thus increases computational requirements. 
\citet{megabyte} employ a mix of multiple models to improve this drawback of byte-wise processing. I.p. they introduce a fixed character-embedding aggregation and a second character-decoder model. However, they use a fixed byte-width that is processed at once, which is not aligned with word splits. 

Consequently, prior research has proposed methods for merging bytes, e.g., through state-space models \cite{wang2024mambabyte}.  However, these approaches still result in performance degradation. 
Finally, linguistically motivated approaches have built tokenizers based on known morphological rules \cite{jabbar2024morphpiece}. However, these methods are usually tailored to specific applications and are usually too costly and error-prone for large, general-purpose models.

\citet{fasttext} in particular discusses how adding subword informations, such as trigrams, enriches the encoding of words and leads to reliable compressions. \citet{hashembed} conduct research on the overloading of different hashfunctions to further improve and compress embedding representations. \citet{hashformer} train BERT-style encoder models based on a different set of hashes on words. \citet{canine} propose a multistage encoding scheme that uses hash functions and convolutions to enhance the BERT-encodings of words.

Another line of work to reduce the vocabulary parameter count is the utilization of weight tying, effectively halving it, as  the embedding and head layers become ``tied'' to the same matrix  \cite{press2017using}. 
However, the effects on performance are still not sufficiently explored, and it arguably imposes a more difficult training objective. 


\section{Conclusion}
\label{sec:conclusion}
In this work we present \method, an alternative to subword tokenizers with a simple and explicitly modeled robust hash function on words. It removes the need and pitfalls to limit ``a models potential'' to a ``pre-pre-trained'' vocabulary.
We, moreover, fundamentally shift the established target of training language models, previously designed as a single-label problem, into a multi-label prediction based on word similarities.
Similarities in particular include leading whitespaces and uppercase variations, for which subword tokenizers add specific tokens that are independently trained from scratch.
These contributions allow us to train language models more robust, more adaptable when continuing pre-training with a new language, and with a significantly (to 12.5\%) reduced parameter size without a decrease in benchmark scores. Due to the special role of the matrices, the latter in particular allows one to increase micro-batchsize, which further accelerates training time.
Finally, the consequent convolution-like encoding achieves SOTA fertility scores across most languages and enables by design synergies to similar language groups.
We demonstrated the latter showing that our $3B$ almost achieved ``native-language'' performance after a small amount of language-transfer training steps, in contrast to the tokenizer baseline.

\section*{Limitations}

With \method\ we propose a fundamentally different approach to text encoding and decoding in LLMs.
Due to the intense resources required to train LLMs, we have focused on evaluating models up to $3B$ parameters. Evaluations on even larger models and training datasets remain a relevant point of investigation for future work. Nonetheless, we observed an easy transfer from $1B$ to $3B$ parameters, and we will continue to train and release more advanced models.

We expect \method\ to experience some numerical instabilities for very long words since single-word embeddings are calculated as the sum of their $n \cdot m$ activations. However, less than $2\%$ of the entire slimpajama dataset contains words with more than $10$ characters (\emph{cf.} App.~\ref{app:statistics}), and we did not encounter any issues with the benchmarks. Consequently, such potential instabilities remain statistically insignificant.
Nonetheless, we could adequately tackle long outliers with an additional split rule based on the words length or at the occurrence of repetitions. Subword tokenizers already demonstrate that such approaches will work, even when tokens are at first glance meaningless and underutilized---and again, these cases remain outliers. Moreover, a hybrid setup utilizing a large tokenizer (>512k tokens) with \method~for optimized memory footprint is depicted in Figure~\ref{fig:hybrid}.

Similarly, we did not thoroughly study the effect of repetitive trigrams in words. These did also not occur frequently enough to have any measurable effect on our experiments. As of now, we only accumulate a word pattern in a binary fashion, not accounting for trigrams appearing multiple times in a single word.
As a fallback, one could again, split words at the position of repetitions. Another promising direction would overload embeddings with positional encodings similar to rotary~\cite{rotary}.

Although \method's fertility on code is on par with that of LLama2 (\emph{cf.} App.~\ref{app:fertility}), it could be further improved by explicitly modeling code patterns. In this work, we have focused on natural language and leave detailed evaluations of \method\ in downstream coding tasks for future research.
Furthermore, we did not investigate languages entirely relying on Unicode byte-encodings, such as Chinese. However, as they seemingly work out-of-the-box with subword tokenizers, we do not expect issues here by splitting them character/ word-wise. In particular for asian symbols, additionally translating the symbols to its romanization through the phonetic alphabet such as pinyin may further improve the synergies of word encodings. 


Finally, we only studied a single constructed hash function for \method. As this work paves the way to model required language features more explicitly, we are looking forward to variations of the proposed \method~method.


\section*{Acknowledgments}


We gratefully acknowledge support by the German Center for Artificial Intelligence (DFKI) project “SAINT”,
the Hessian Ministry of Higher Education, the Research and the Arts (HMWK) cluster projects
“The Adaptive Mind” and “The Third Wave of AI”, and the ICT-48 Network of AI Research Excellence Center “TAILOR” (EU Horizon 2020, GA No 952215). 



\bibliography{references}

\clearpage

\appendix

\section*{Appendix}

\section{\method\ Algorithm}
\label{app:mesa_algo}
Alg.~\ref{alg:split},\ref{alg:mesa_algo},\ref{alg:encode},\ref{alg:dict},\ref{alg:decode} show the core steps to encode text into embeddings, and decode text from model predictions with \method.
Here, \emph{regex.split} denotes an algorithm that splits text based on a regular expression, \emph{hash} denotes an arbitrary hash function like \emph{md5}, $\%$ denotes the mathematical \texttt{modulo} operation.
In style of python, $f'\{token\}\_'$ denotes text formatting to indicate the string with content of variable \emph{token} being followed by an underscore, and $EL[i]$ denotes the $i-$th entry of matrix $EL$ and $'string'[i:i+3]$ three consecutive characters in the text \emph{string} starting from position $i$, where $'s'$ is at position $0$.
Finally, $v\approx 8,000$ is the chosen vocabulary size,
$d\approx 100,000$ is the chosen dictionary size, $h\approx 3,072$ the LLMs hidden size. 
Finally, $\mathbb{0}^h$ denotes a zero vector of dimension $h$ and $\mathbb{1}^{v\times d}$ a matrix with entries $0$ or $1$.
Note that we included some normalization steps in Alg.~\ref{alg:decode}, which we surprisingly found not beneficial for Alg.~\ref{alg:encode} in our ablations.

\begin{figure*}
    \centering
\includegraphics[width=\linewidth,clip,trim=0 0 4cm 0]{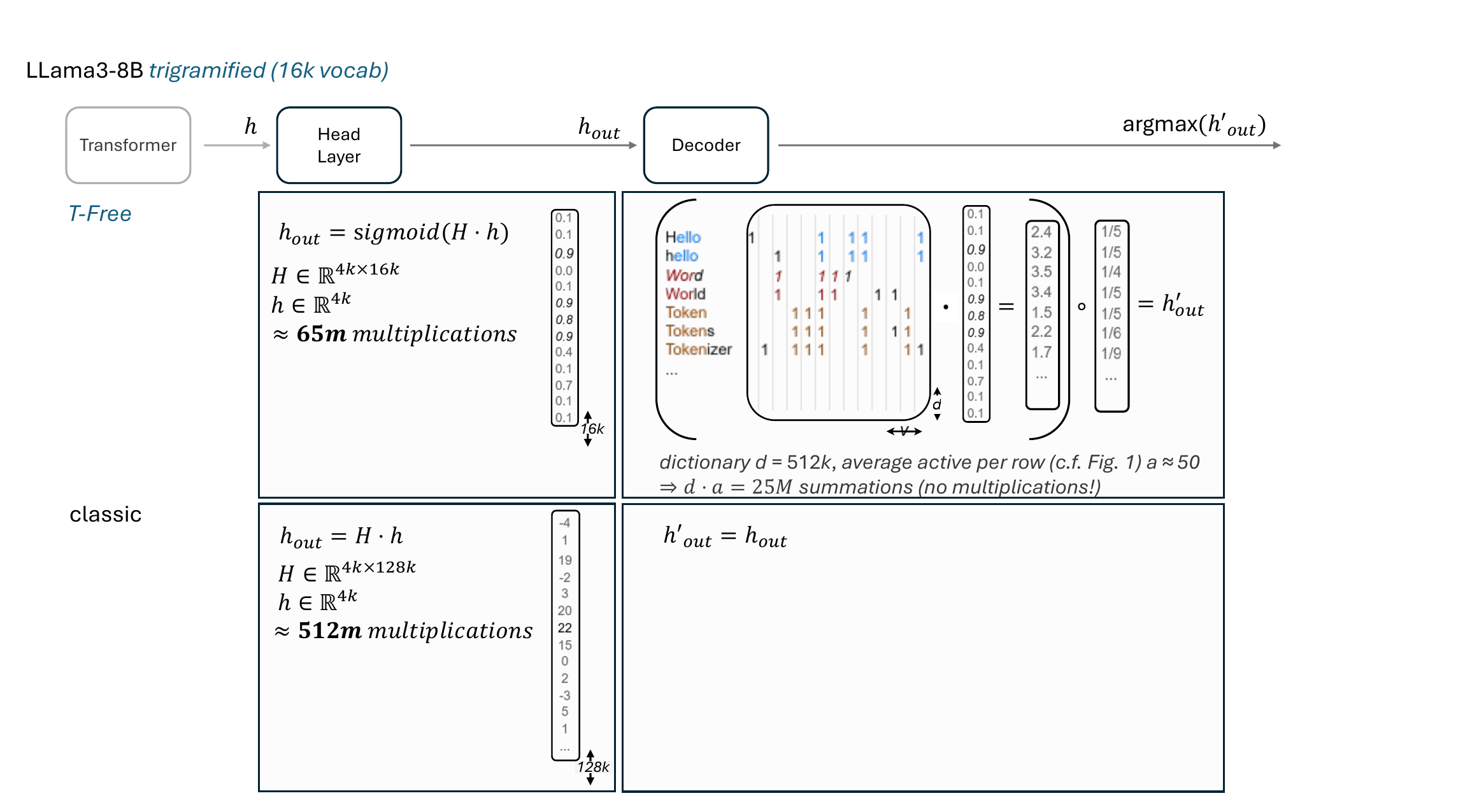}
    \caption{``Greedy'' text-decoding example for T-Free (top) and classic decoder LLMs (bottom). T-Free applies a head of significantly reduced parameters which results in less dense matrix multiplications and smaller vector sizes. As an additional step, during inference, T-Free computes the \emph{average activation score} $h'_{out}$, which is sparsely computed by multiplying (and averaging) the once precomputed decodable dictionary with the sigmoid scores of the head. Finally, in both cases argmax is taken to lookup the resulting word.}
    \label{fig:decoder}
\end{figure*}

\begin{figure*}
    \centering
\includegraphics[width=\linewidth,clip,trim=0 2.5cm 0 0]{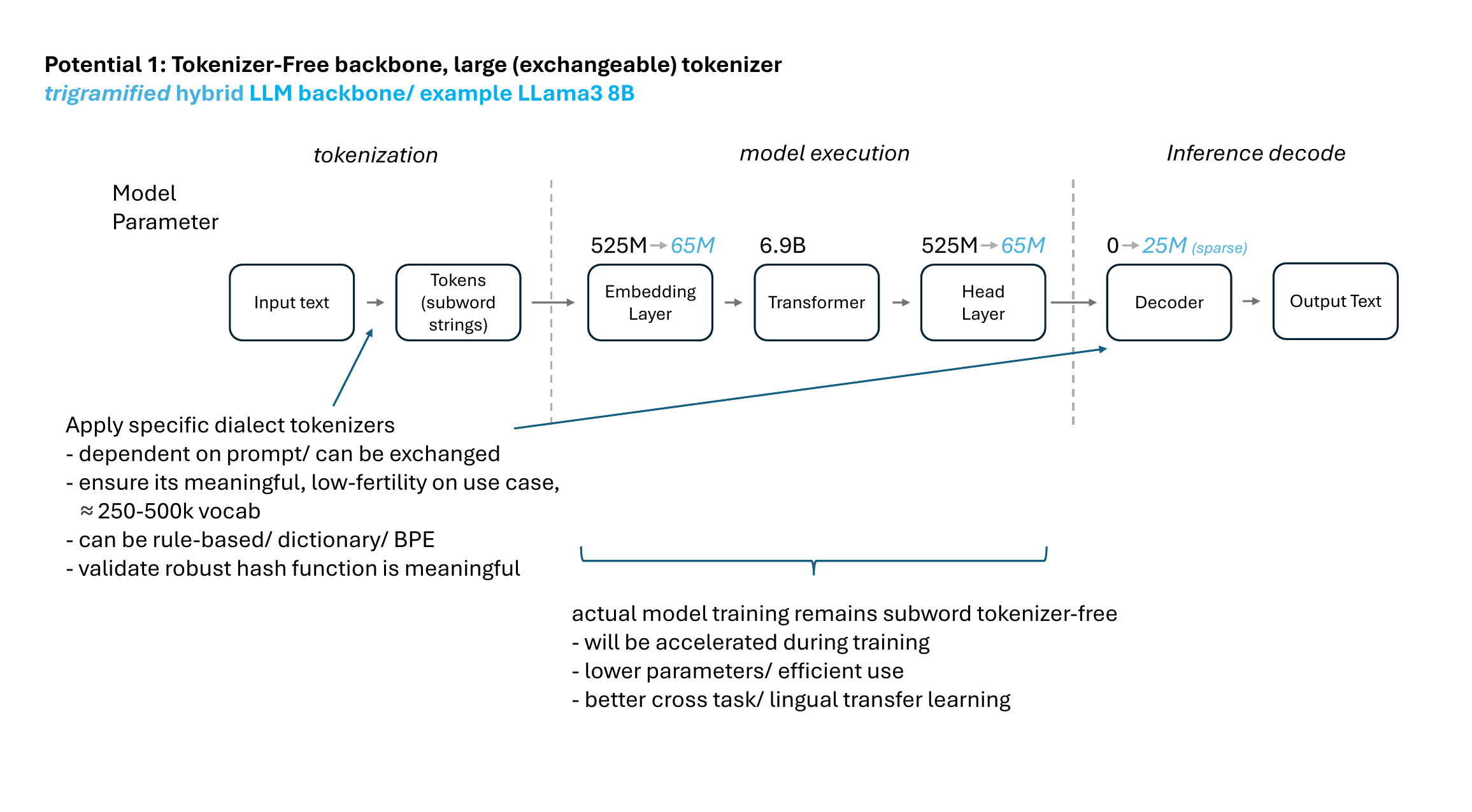}
    \caption{Hybrid ``T-Free'' (tokenizer-free/adaptable) LLM Backbone applying large scale (500k+) tokenizers. Major advantages of a T-Free backbone in a hybrid setting are the compression of embedding and head matrices, and the potential flexibility to lateron exchange (with some finetuning) the tokenizer --- the backbone remains with the same tokenizer-free encoding rules.}
    \label{fig:hybrid}
\end{figure*}

Finally, refer to Figure.~\ref{fig:llm_pipeline},\ref{fig:decoder} for a step-wise comparison of the computation step, parameters and runtimes. Figure~\ref{fig:hybrid} shows a ``hybrid'' mode, in which embody a classical subword-tokenizer as a text preprocessing step, but utilize \method~to keep the ``tokenizer free LLM backbone''. Arguably, this approach benefits from a compressed embedding layer, and the tokenizer may easier be exchanged afterwards---the encoding of the text-chunks in the backbone will be kept as proposed.

\begin{algorithm}[h]
\caption{token\_split}\label{alg:split}\label{alg:mesa_algo}\label{alg:mesa_algo_decode}
\begin{algorithmic}
\State \textbf{input:} \emph{text} \vspace{4px}
\State $\textrm{tokens}\gets \textrm{regex.split}((\_|\backslash W|\backslash d), text)$
\State \emph{(cf. Sec.~\ref{sec:whitespace} if necessary)}\vspace{4px}
\State \textbf{output:} \emph{tokens}
\end{algorithmic}
\end{algorithm}

\begin{algorithm}[h]
\caption{trigramify}\label{alg:trigramify}\label{alg:mesa_algo}\label{alg:mesa_algo_decode}
\begin{algorithmic}
\State \textbf{input:} \emph{token, k, m} \\
\Comment{$k:$ lowercase activation, $m:$ total activation} \vspace{4px}
\State $pattern \gets \mathbb{0}^{v}$
\For{$l\in [0,len(token)-1]$}
\State $trigram \gets f'\_\{token\}\_'[l:l+3]$
\For{$i \in [1,m]$}
\If{$i \leq k$}
\State $string_i = \textrm{lower}(trigram)$
\Else
\State $string_i = trigram$
\EndIf
\State $hash_i = \textrm{hash}(f'\{string_i\}\_\{i\}')$
\State $pattern[hash_i\%v] = 1$
\EndFor\vspace{4px}
\EndFor\vspace{4px}
\State \textbf{output:} \emph{pattern}
\end{algorithmic}
\end{algorithm}

\begin{algorithm}[h]
\caption{encode}\label{alg:encode}
\begin{algorithmic}
\State \textbf{input:} \emph{token}, \emph{EL}\\
\Comment{\emph{EL}: Embedding Layer ($\in \mathbb{R}^{v \times h}$)\phantom{abdcdef}}\vspace{4px}
\State $embedding \gets \mathbb{0}^{h}$
\State $pattern \gets \textrm{trigramify}(token)$
\For{$i\in [0,v-1]$}
\If{$pattern[i] == 1$}
\State $embedding \gets embedding+EL[i]$
\EndIf
\EndFor
\State \textbf{output:} \emph{embedding}
\end{algorithmic}
\end{algorithm}

\begin{algorithm}[h]
\caption{compile\_dictionary}\label{alg:dict}
\begin{algorithmic}
\State \textbf{input:} \emph{tokens}\vspace{4px}
\Comment{$d$ target tokens} 
\State $dict \gets \mathbb{0}^{d\times v}$
\For{$i\in [0,d-1]$}
\State $dict[i] \gets \textrm{trigramify}(tokens[i])$
\EndFor
\State \textbf{output:} \emph{dict}
\end{algorithmic}
\end{algorithm}

\begin{algorithm}[h!]
\caption{decode}\label{alg:decode}
\begin{algorithmic}
\State \textbf{input:} \emph{logit}, \emph{dict}, \emph{tokens}\\
\Comment{\emph{logit}: single prediction ($\in \mathbb{R}^{v\times 1}$),\phantom{abcdefg}\\
\phantom{>bcdef}\emph{dict}: compiled dictionary ($\in \mathbb{1}^{d\times v}$),\\
\phantom{>bcdef}\emph{tokens}: $d$ tokens corresponding to \emph{dict} }\vspace{4px}
\State $scores \gets dict \cdot \textrm{sigmoid}(logit)$
\For{$i\in [0,d-1]$}
\State $scores[i] \gets scores[i]/\textrm{sum}(dict[i])$
\EndFor
\State $scores \gets \textrm{softmax}(scores)$
\State $i \gets \arg\max_l\ scores[l]$\vspace{4px}
\State \textbf{output:} \emph{tokens[i]}, \emph{scores[i]}
\end{algorithmic}
\end{algorithm}

\section{Whitespace encoding}
\label{sec:whitespace}
By default our model is trained to predict full words separated by whitespaces. To not be limited to this use-case, we add a special ``non-whitespace'' and ``whitespace'' token. We empirically evaluated each exception occuring in code tokenization.
To further reduce its fertility, we favor ``non-whitespace'' before one of the following characters:
\begin{lstlisting}
$.,;:#?!=-+*/\()<>[]&@%_~^
\end{lstlisting}
We further prefer non-whitespace after one of the following characters:
\begin{lstlisting}
#$=-+*/'\"(<[~^&@%_\n1234567890
\end{lstlisting}

As such, the text ``In 2024'' would result in the split ``[In,2,0,2,4]'' without the need of any special annotations, while ``In20 24'' resolves to ``[In,<no\_ws>,2,0,<ws>,2,4]''.

Finally, to further improve code fertility, we merge consecutive <ws> and newline tokens up to 3 times, i.e. $8$ consecutive whitespaces would result in a single <|8<ws>|> token.

\section{Tokenizer trainings with sentencepiece}
\label{app:sentencepiece}

For training of a unigram tokenizer with the current sentencepiece library, a 20GB reference data corpus reaches the limit of our available 1TB Ram compute node.
We thus randomly sample 20GB of the slimpajama dataset and run the following statement for training of the actual tokenizer: 
\begin{lstlisting}
spm_train --input=20GB_sample.txt\
--model_prefix=unigram_64k \
--vocab_size=64000 \
--character_coverage=0.99 \
--model_type=unigram \
--byte_fallback=true \
--split_by_number=true  \
--split_by_whitespace=true  \
--train_extremely_large_corpus=true\
--split_digits=true  \
--allow_whitespace_only_pieces=true\
--remove_extra_whitespaces=false \
--normalization_rule_name=nfkc \
--num_threads 64 --eos_id=0 \
--bos_id=-1   --unk_id=2 \
--pad_id=1 \
--eos_piece="<|endoftext|>" \
--pad_piece="<|padding|>" \
--unk_piece="<|unknown|>"
\end{lstlisting}

\section{Training Configurations}
\label{app:training_parameters}
\subsection{1B}
Training Parameters are listed in Tab.~\ref{tab:1b_parameters}.

\begin{table}[]
    \centering
\begin{tabular}{lrr}
\toprule
         Parameter & Value \\
\midrule
hidden size & 2,048 \\
layers &  16 \\
attention heads & 16 \\
norm & layer \\
mlp & gelu \\
mlp scale & 5,456 \\
training steps & 50k \\
sequence length & 2,048 \\
batch size & 1,024 \\
\midrule
precision & bfloat16 \\
learning rate &  6e-4 \\
minimum learning rate & 6e-5 \\
annealing & cosine \\
annealing steps & 50k \\
warmup steps & 200 \\
optimizer & AdamW \\
optimizer beta1/ beta2/ eps & 0.9 / 0.95 / 1e-8 \\
weight decay & 0.1 \\
\bottomrule
    \end{tabular}
    \caption{1B Parameter configurations (for all ablations).}
    \label{tab:1b_parameters}
\end{table}

\subsection{3B}
Training Parameters are listed in Tab.~\ref{tab:3b_parameters}.

\begin{table}[]
    \centering
\begin{tabular}{lrr}
\toprule
         Parameter & Value \\
\midrule
hidden size & 3,072 \\
layers &  24 \\
attention heads & 24 \\
norm & rms \\
mlp & swilu \\
mlp scale & 8,192 \\
training steps & 90k (20k) \\
sequence length & 4,096 \\
batch size & 1,024 \\
\midrule
precision & bfloat16 \\
learning rate &  3e-4 (1e-4) \\
minimum learning rate & 3e-5 (3e-5)\\
annealing & cosine \\
annealing steps & 90k (20k) \\
warmup steps & 200 (500) \\
optimizer & AdamW \\
optimizer beta1/ beta2/ eps & 0.9 / 0.95 / 1e-8 \\
weight decay & 0.1 \\
\bottomrule
    \end{tabular}
    \caption{3B Parameter configurations (for all ablations). In brackets are highlighted values for German continued pre-training.}
    \label{tab:3b_parameters}
\end{table}

\section{Fertility Analysis}
\label{app:fertility}
We subsequently provide further experimental details on the fertility analysis conducted with respect to \textbf{F3}, Sec.~\ref{sec:main_results}. 
As a reference dataset, we used the November 23 dump of Wikipedia in the respective languages. We derived reference tokenization using UDPipe \cite{straka-2018-udpipe}. A tokenizer's fertility is then calculated by dividing its total token count for a document by the number of tokens produced by UDPipe.
We present results for more models on 8 languages in Tab.~\ref{tab:language_fertility_complete}.

We also evaluated the white-space tokenization of \method\ for code. For 22 programming languages, we took 10k random documents each from the starcoder dataset \footnote{\url{https://huggingface.co/datasets/bigcode/starcoderdata}}. Since ground truth text splitting for code is hard to establish, we instead report the normalized sequence length with respect to a reference tokenizer. We here used Llama-2 and report results in
Tab.~\ref{tab:code_nsl}. Since \method's tokenization achieves an NSL close to $1.0$, it performs roughly on par with Llama-2.
\section{Token Overlap/Duplicates}
For the empirical evaluation regarding \textbf{F2}, \emph{cf.} Sec.~\ref{sec:exp_flaws}, we present more exhaustive results with additional models in 
Tab.~\ref{tab:token_overlaps}.
 
\begin{table}[]
    \centering
\begin{tabular}{lrr}
\toprule
     lang & Ours (NSL) $\downarrow$ &  Starcoder (NSL)  $\downarrow$\\
\midrule
    c-sharp &  1.034783 &   0.816206 \\
          c &  0.996308 &   0.860453 \\
        cpp &  1.084867 &   0.855094 \\
        css &  1.109492 &   0.903693 \\
       cuda &  1.018222 &   0.857034 \\
 dockerfile &  0.954086 &   0.851568 \\
         go &  1.142476 &   0.883456 \\
       html &  1.164936 &   0.885237 \\
       java &  1.003201 &   0.835858 \\
 javascript &  1.183923 &   0.850398 \\
       json &  1.071685 &   0.892871 \\
     kotlin &  0.925868 &   0.846053 \\
   makefile &  1.006108 &   0.862994 \\
   markdown &  0.965325 &   0.892784 \\
        php &  1.179374 &   0.838566 \\
     python &  1.005064 &   0.857439 \\
       ruby &  0.979135 &   0.846597 \\
       rust &  1.086027 &   0.857645 \\
      shell &  1.041322 &   0.879112 \\
        sql &  0.954786 &   0.859785 \\
 typescript &  1.121119 &   0.847393 \\
       yaml &  0.974146 &   0.856218 \\

\textbf{Overall} & 1.045557 & 0.860748 \\
\bottomrule
\end{tabular}
    \caption{Normalized sequence length wrt Llama-2 on code tokenization.}
\label{tab:code_nsl}
\end{table}

\begin{table*}[]
\small
    \centering
    \begin{tabular}{l | r r r r r r r r}
     \textbf{Model}    &  \textbf{EN} & \textbf{DE} & \textbf{FR} & \textbf{ES} & \textbf{IT} & \textbf{RU} &\textbf{VI} &\textbf{AR} \\
     \hline
      
      Unigram Baseline (32k) & 1.3584 & 2.2577  & 2.1354 & 2.1524 & 1.9508  & 11.4448 & 5.1826 & 9.4740\\
      Unigram Baseline (64k) & 1.2802 & 2.0043 & 1.9492 & 1.9163  & 1.7263 & 11.4305& 5.0603 & 9.4555\\
      BPE Baseline (32k) & 1.3585 & 2.2784  & 2.0625 & 2.0977 & 1.9396 & 11.4321 & 4.8717 & 9.4694\\
      BPE Baseline (64k) & 1.2759 & 2.0253 & 1.9059 & 1.8894 & 1.7212 & 11.4231 & 4.7545 & 9.4656\\
      Mistral (32k) &  1.3973 & 1.9316 & 1.6613 & 1.7569 & 1.7591 &  2.5601 & 3.3458 & 4.7228\\
      Llama-2 (32k) & 1.4014 & 1.7702 & 1.5495 & 1.6413 & 1.6160 & 2.3242 & 3.3540 & 4.8255\\
      Phi-2: (50k) & 1.2654  & 2.2660 & 1.8183 & 1.9736 & 1.9132 & 6.7289 & 4.3392 & 5.2246\\
      Gemma (256k) & 1.1761 & 1.4470 & 1.2754 & 1.3163 & 1.3253 & 1.9028 & 1.7257 & 1.7938\\
      DBRX (100k) & 1.2381 & 1.8311 & 1.5423 & 1.6142 & 1.6191 & 3.2385 & 2.6617 & 3.6821\\
      Jais (85k) & 1.3029 & 2.1391 & 1.7347 & 1.8514 &  1.8244 & 3.6730 & 3.4382 & 1.2653 \\
      Command-R (255k) & \textbf{1.1525$\bullet$} & 1.4110 & 1.2079 & 1.2527 & 1.2460 & 1.5899 & 1.5967 & 1.5787\\
      Llama-3 (128k) & 1.2330 & 1.8221 & 1.5351 & 1.6033 & 1.6130 & 2.2144 & 1.8261 & 1.9660\\
      NeMo-Tekken (131k) & 1.2313 & 1.5178 & 1.3061 & 1.3845 &  1.4171 & 2.0521 & 1.8378 & 1.6045 \\
      \textbf{Ours}   &  \textbf{1.1636$\circ$} & \textbf{1.1829} & \textbf{1.2363} & \textbf{1.1695} & \textbf{1.1274} & \textbf{1.3386} & \textbf{1.4001} & \textbf{1.0863}\\

    \end{tabular}
    \caption{Additional evaluations of fertility evaluations. \emph{Cf.} Sec.~\ref{sec:exp_flaws}.}
    \label{tab:language_fertility_complete}
\end{table*}

\begin{table*}[]
     \small
    \centering
    \begin{tabular}{l | c r r r|}

    \multirow{2}{*}{\textbf{Model/Tokenizer}} & \multicolumn{4}{ c |}{\textbf{Portion of duplicate tokens (\%) $\downarrow$}}\\
         &  \textbf{Total} &\textbf{Cap.}  &  \textbf{Space} & \textbf{Digits}\\
     \hline
      Unigram Baseline (32k) & 32.99 & 21.44 & 11.76 & 0.00\\
      Unigram Baseline (64k) & 35.24 & 23.27 & 13.47  & 0.00\\
      BPE Baseline (32k) & 32.12 & 21.30 & 13.85 & 0.00\\
      BPE Baseline (64k) & 35.32 & 23.82 & 15.52  & 0.00\\
      Phi-2: (50k) & 23.23 & 12.91 & 16.89 & 3.32\\
      DBRX (100k) & 24.87 & 23.77 & 16.17 & 1.10\\
      GPT-2 (50k) & 25.25 & 21.93 & 16.99 & 3.32\\
      Gemma (256k) & 34.68 & 20.27 & 20.50 & 0.04\\
      Command-R (255k) & 15.31 & 15.31 & 14.00 & 0.00\\
      Mistral (32k) & 31.47 & 19.10 & 16.45 & 0.00\\
      Llama-2 (32k) & 30.23 & 17.10 & 16.98 & 0.00\\
      Llama-3 (128k) & 21.22 & 20.17 & 15.28 & 1.05\\
      NeMo-Tekken (131k) & 23.12 & 13.30 & 11.99 & 0.00\\
      \textbf{T-Free} (Ours) & \textbf{0} & \textbf{0} & \textbf{0} & \textbf{0}\\
    \end{tabular}
    \caption{Additional evaluations of overlap of full tokens occuring multiple times, only with capitalization or whitespace in difference. Note that there are still plenty more redundancies with sub-token reconstructions. \emph{Cf.} Sec.~\ref{sec:exp_flaws}.}

    \label{tab:token_overlaps}
\end{table*}

\section{Training stability}
\label{app:training_stability}
\begin{figure}
    \centering
    \begin{subfigure}{\linewidth}
    \includegraphics[width=\linewidth]{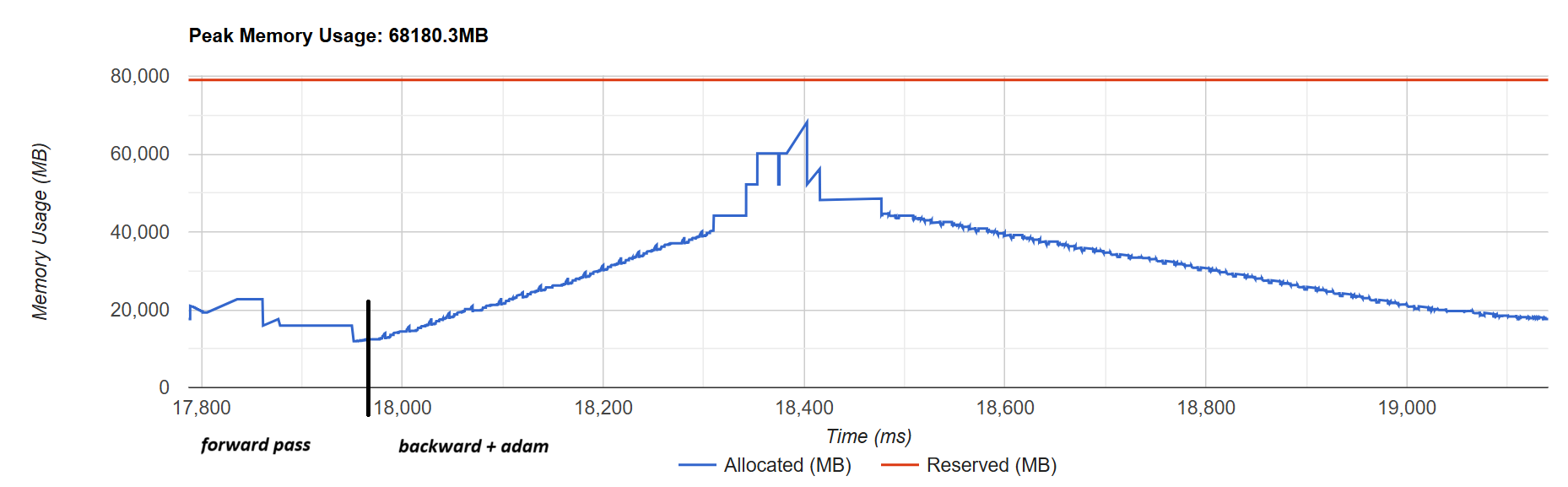}
    \caption{Classic Tokenizer.}
    \end{subfigure}
    \begin{subfigure}{\linewidth}
    \includegraphics[width=\linewidth]{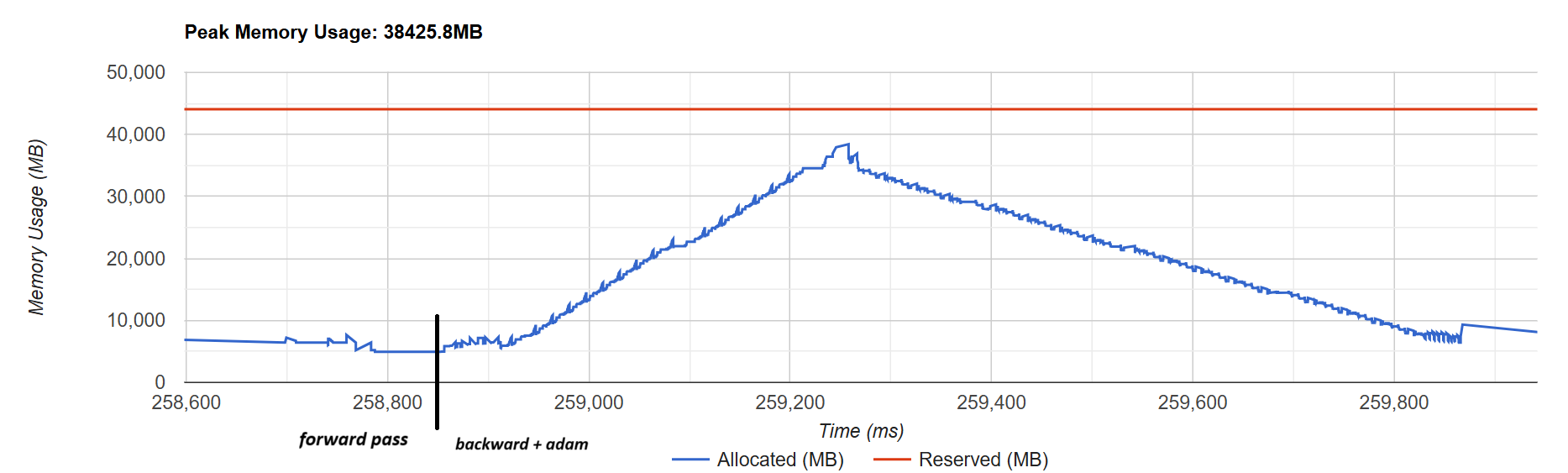}
    \caption{\method.}
    \end{subfigure}
    \caption{Pytorch Profiler Memory Footprint of a single forward and backward pass on a $1B$, each with batch size $8$ and $4k$ sequence length. Top is classical tokenizer version with $64k$ vocab size, bottom trigram with $8k$ vocabulary. Note how AdamW aggregates peak memory consumption until 68GB for classic tokenizer, while ours remains at 38GB.}
    \label{fig:memory_footprint}
\end{figure}

\begin{figure*}
    \centering
\includegraphics[width=\linewidth,clip,trim=0 7cm 0 0]{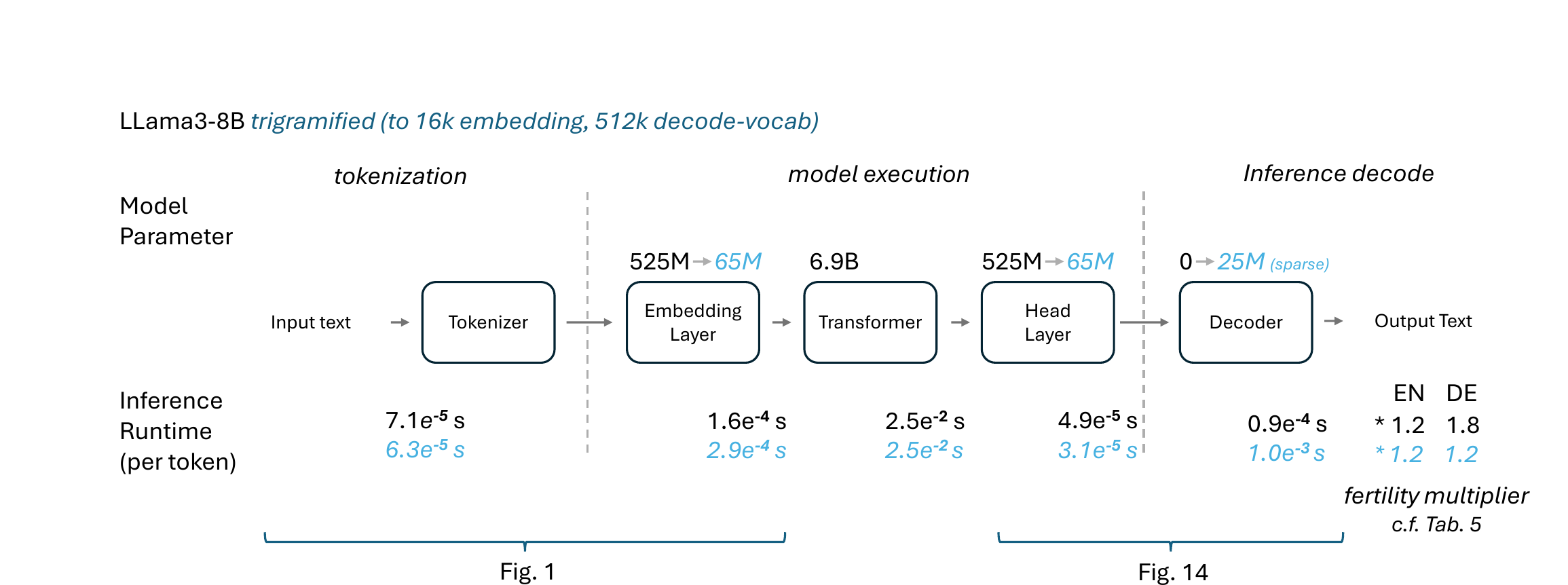}
    \caption{Comparison of the end-to-end LLM processing steps for the standard LLama3-8B model versus a proposed \emph{trigramified} version. In particular, the two biggest matrices of the model, the embedding layer and the head can be significantly compressed, which can half the training resources when using standard libraries (\emph{c.f.} Fig.~\ref{fig:memory_footprint}). Otherwise, the training execution time is mostly on par. 
    For decoding, the proposed T-Free version requires an additional step to predict the next word. We assumed a vocabulary of 512$k$ entries with an average of 50 activations per entry. This leads to additional 25M nonzero parameters that can be casted into a sparse matrix format (\emph{c.f.} Fig.~\ref{fig:decoder}). The overall inference run-time increases slightly when averaging the entire pipeline processing time, but the biggest consumption remains at the actual transformer backbone. However, note that in addition training and inference time benefit from the improved fertility factors of T-Free. Furthermore, depending on the use case, smaller dictionaries, faster sparse hardware accelerators, or different decoding strategies may be applicable.}
    \label{fig:llm_pipeline}
\end{figure*}

Memory footage comparing classic tokenizers to \method\ is found in Fig.~\ref{fig:memory_footprint}.

\begin{figure*}
    \centering
    \begin{subfigure}{\linewidth}
    \includegraphics[width=\linewidth]{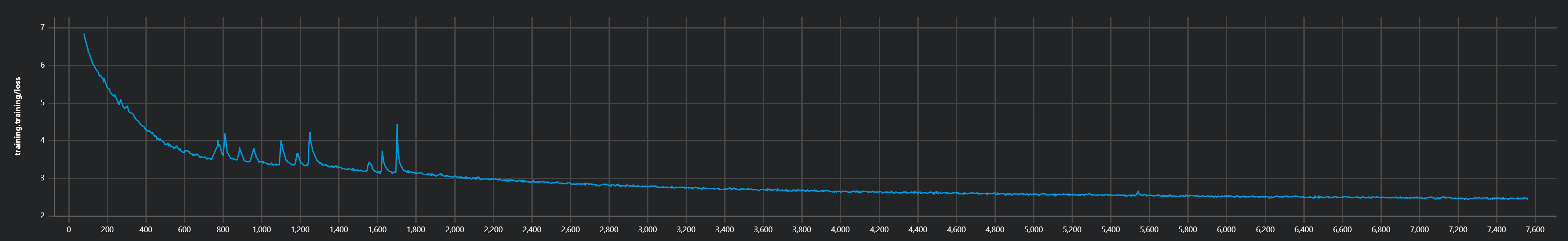}
    \caption{Classic Tokenizer.}
    \end{subfigure}
    \begin{subfigure}{\linewidth}
    \includegraphics[width=\linewidth]{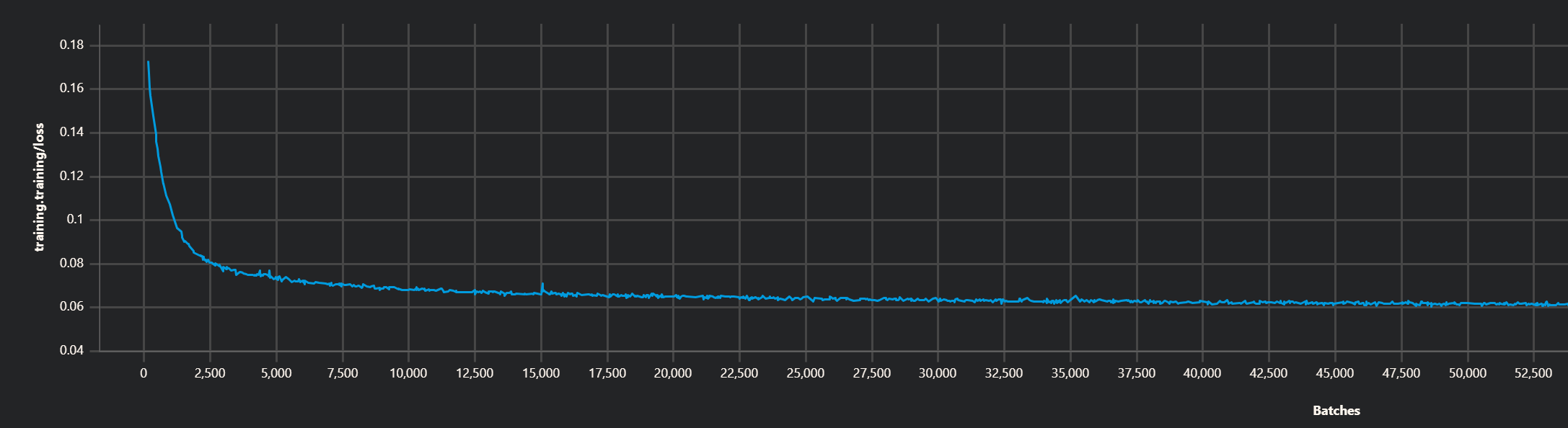}
    \caption{\method.}
    \end{subfigure}
    \caption{Exemplary comparison of classic tokenizer ($v=64k$) training  loss curve (top) and \method\ ($v=16k$) training loss (bottom). Overall we noticed less spikey training behavior when using \method. Both 3B models were trained on same slimpajama data, token-batchsize and learning rate 4.5e-4.}
    \label{fig:training_loss}
\end{figure*}

Note that the hashing step of Alg.~\ref{alg:trigramify} uniformly distributes gradients amongst the available vocabulary, as discussed in Sec.~\ref{sec:discussion}.
This is in contrast to classic tokenizers, as they depend on a bijective single-label mapping, and as such each vocabulary entry update is dependent on its the occurance frequency of the corresponding token within the dataset.
Moreover, we explicitly let trigram activations overlap with their lowercase version. We assume that these are responsible for the more stable training dynamics as shown in Fig.~\ref{fig:training_loss}.
Moreover, we found that the lowercase overlap bootstraps learning as shown with the downstream benchmark ablations Fig.~\ref{fig:hyperparameter_ablation}.

\section{Hyperparameter Ablations}
\emph{Some 1,500 determined experiments later...}

\begin{figure}
    \centering    
         \includegraphics[width=.95\linewidth]{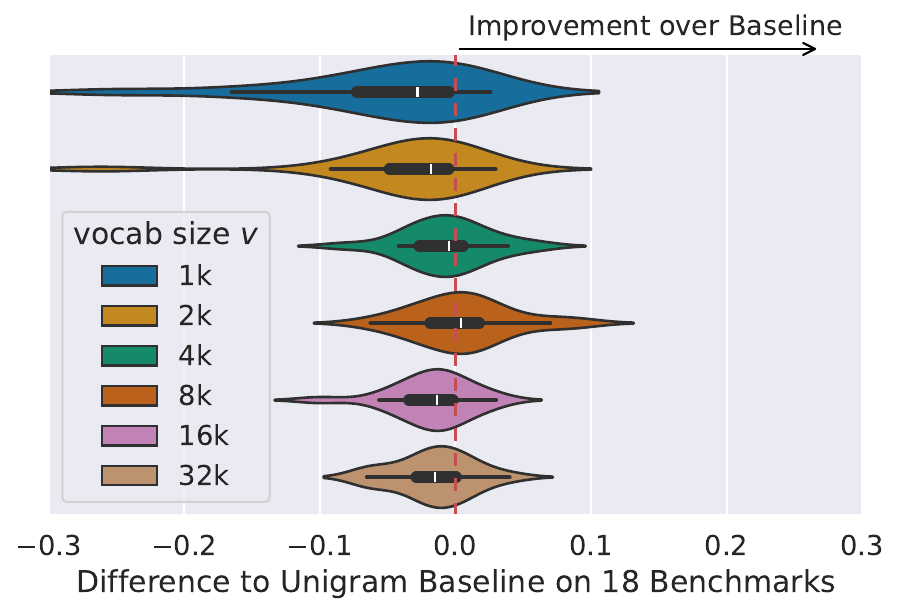}
         \caption{Further ablations on hyper paramters in accordance to Fig.~\ref{fig:hyperparameter_1b}. Note that after peaking at $v=8k$, performance slightly decreases again, which may be attributed to the undertrained stage of the model trainings.}
    \label{fig:hyperparameter_hs}
\end{figure}

\begin{figure}
    \centering    
         \includegraphics[width=.95\linewidth]{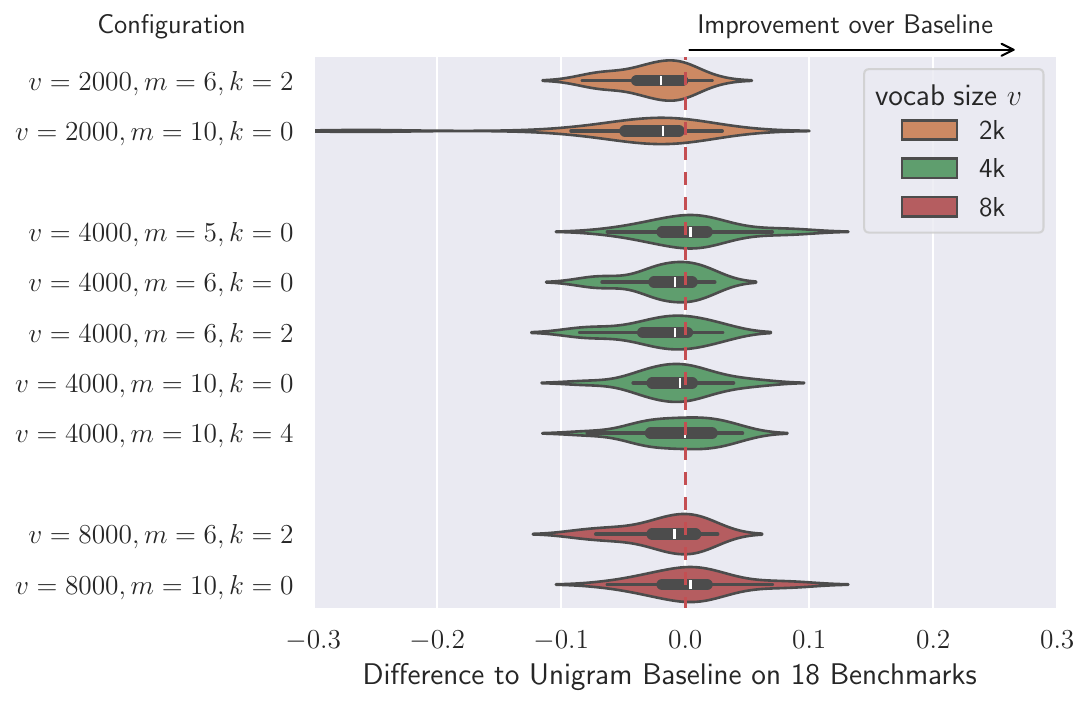}
         \caption{Further ablations on hyper paramters.}
    \label{fig:hyperparameter_ablation}
\end{figure}

\label{app:hyperparameter_ablation}
Albeit pretty scarse, some more hyper-parameter ablations are found in Fig.~\ref{fig:hyperparameter_hs},\ref{fig:hyperparameter_ablation}.

We will continue to polish and add more...

\section{Some Statistics}
\label{app:statistics}

\paragraph{Trigram combinatorics.}
As there are more than $v$ possible words, there will naturally be some overlap in the activations between words. However, assuming an embedding dimension of $v \approx 8,000$, $m \approx 8$ activations per trigram, and a word of length $n = 5$, there are (in theory) ${v \choose n \cdot m} \approx 10^{108}$ unique activation patterns. 

This overlap can be interpreted as an interpolation between input states. For entirely independent inputs, this overlap should be kept small as the results cannot benefit from the states of the shared activations.
As such, we require a \emph{robust hash function} on text, i.e. a mapping from text into sparse activation patterns, for which the overlapping of activations is proportional to the similarity of the input words.
We model this through trigrams, and as such, letter-similarity.

\paragraph{Tokenizer Duplicates.}
Tab.~\ref{tab:word_prefixes} shows the curse of token-based vocabularies: to produce all 64 upper and whitespace variations of the word ``\_words'', one requires on average 3 tokens per writing.

\begin{figure*}
    \centering
    \includegraphics[width=\linewidth]{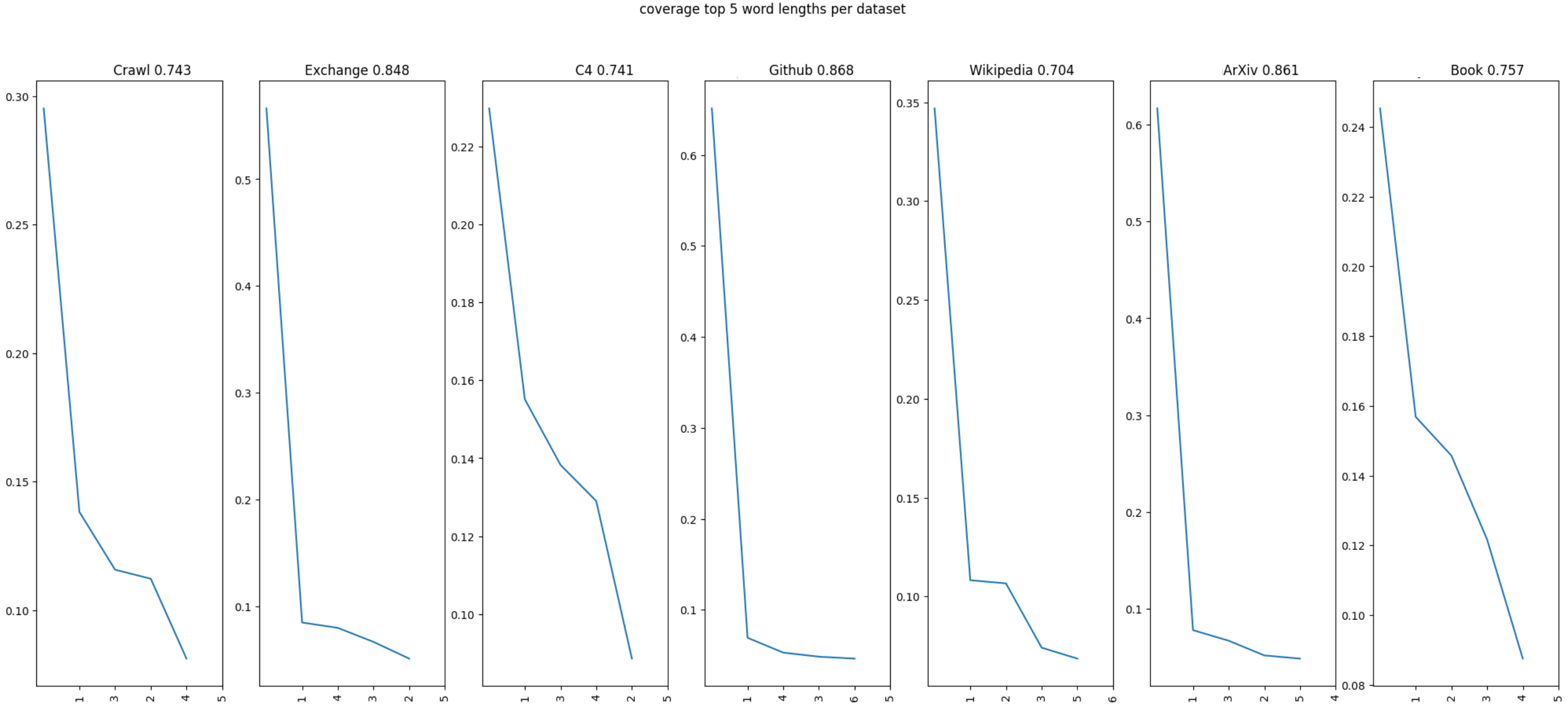}
    \includegraphics[width=\linewidth]{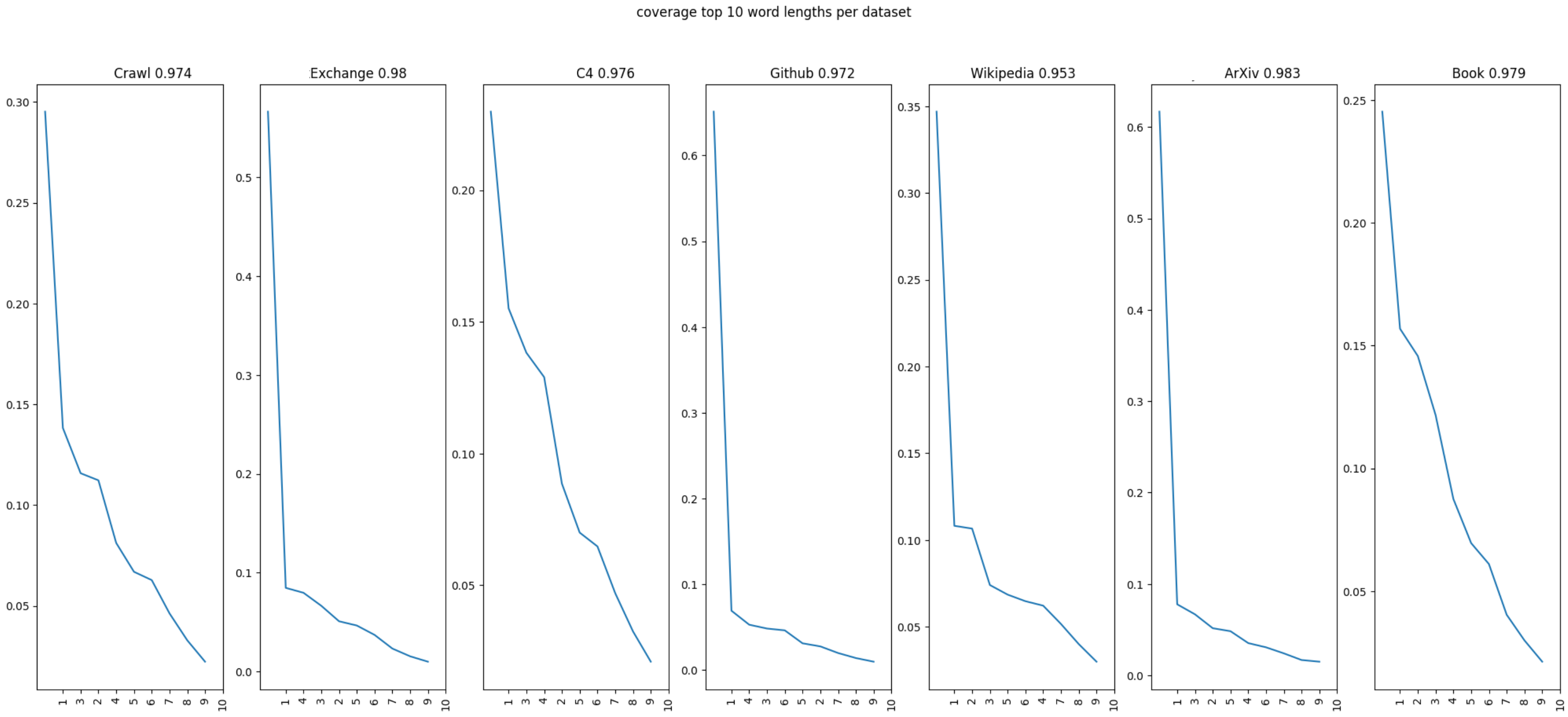}
    \caption{Top 5 and top 10 occuring word lengths (x-axis) per slimpajama data category, with coverage-percentage (y-axis).  Headline indicates total percentage covered by top $n$-length words. With words of length $\leq 5$, one always covers $\geq 74\%$ of all occuring words. With all words of length $\leq 10$, one achieves $\geq 95\%$.}
    \label{fig:top-nwords}
\end{figure*}

\paragraph{Dataset-Coverages.}
Fig.~\ref{fig:top-nwords} shows the covered percentages of the entire dataset, by word-lengths, for all slimpajama datasets.
If we successfully can encode all words of length $\leq 10$, we can cover $\geq 95\%$ of the entire slimpajama dataset. Or conversely, we would only require 5\% outlier handling/ additional splits for longer words (\emph{cf.} Sec.~\ref{sec:conclusion}).

\begin{figure*}
    \centering
\includegraphics[width=\linewidth]{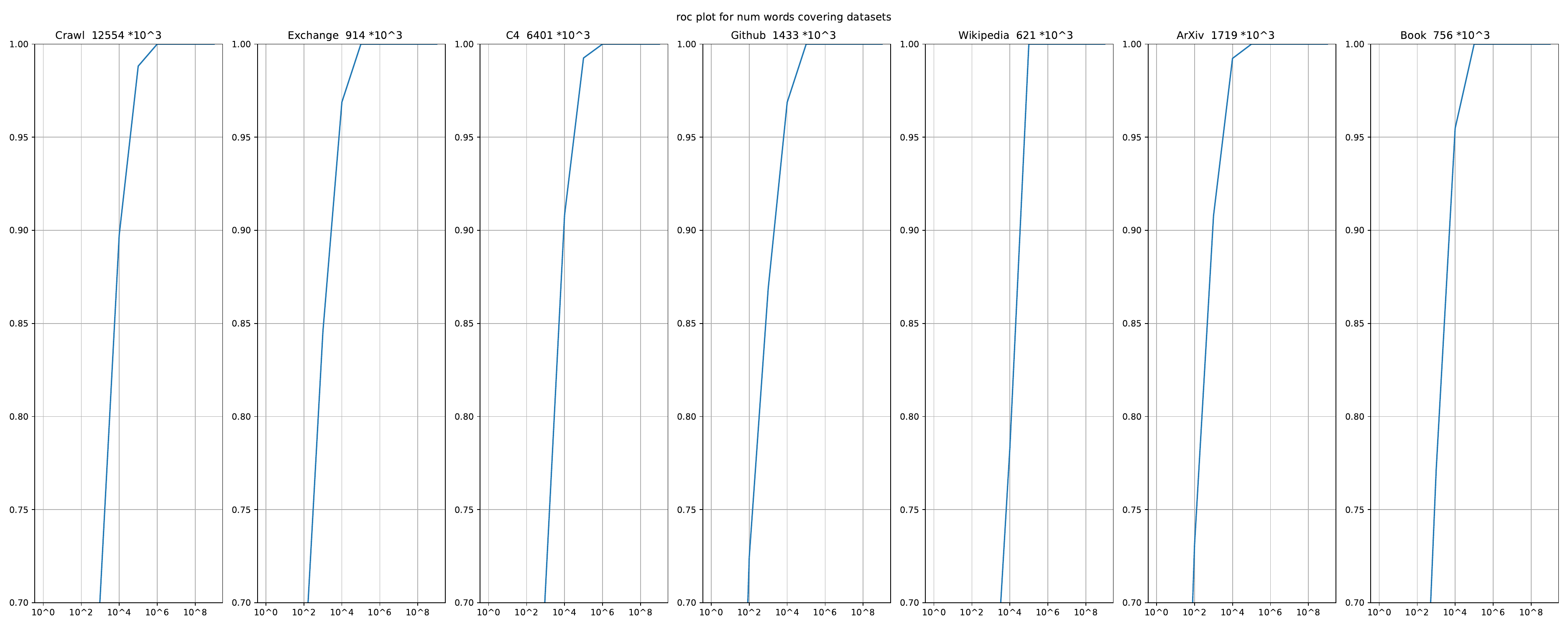}
\caption{Most-frequent word coverage of Slimpajama categories.  Title shows the total number of words per dataset sample, $x$-axis the top-n chosen words, $y$-axis the percentage covered within dataset. With only 100k words we can consistenly cover $> 95\%$ of each category.}
\label{fig:top_100k}
\end{figure*}

\begin{figure*}
    \centering
    \includegraphics[width=\linewidth]{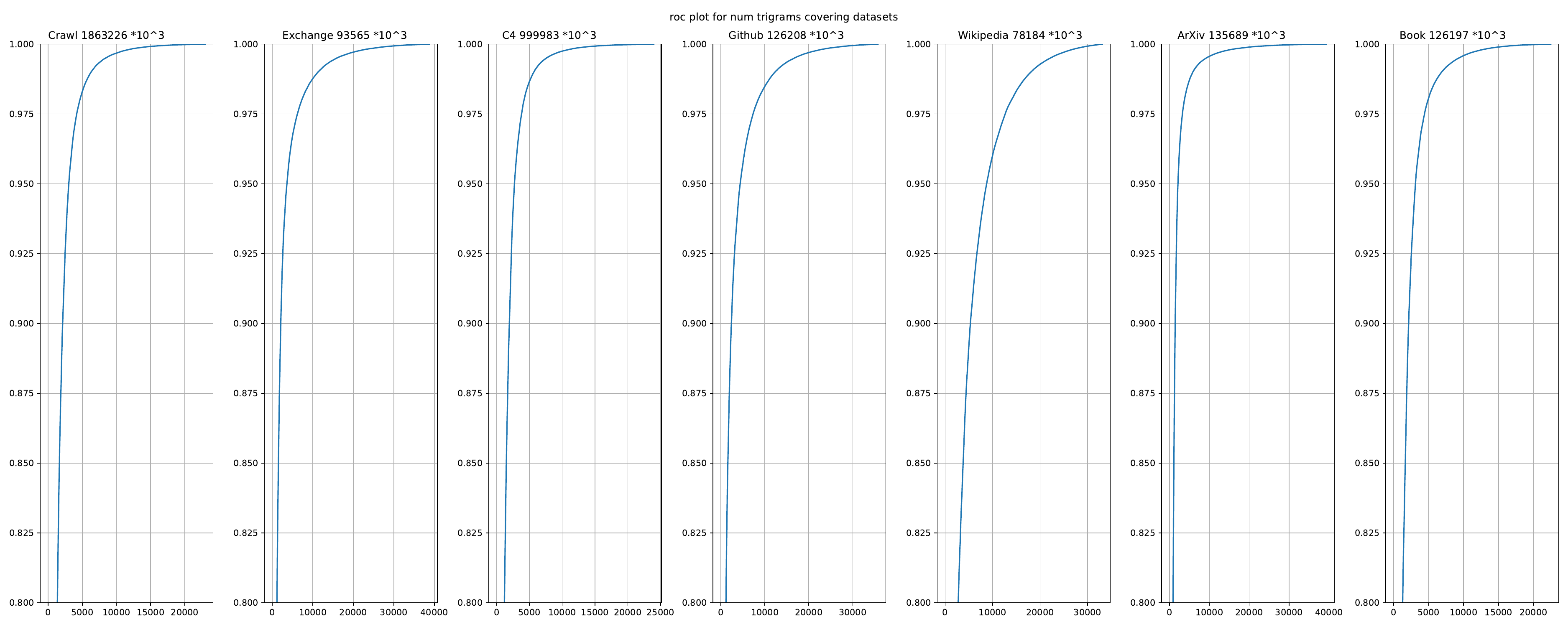}
    \caption{Number of trigrams ($x$-axis) required to cover ($y$-axis) percentage of the categories of slimpajama (total number words sampled in title). With only 10k trigrams we can cover $> 95\%$ of all occuring words.}
    \label{fig:trigrams}
\end{figure*}

Fig.~\ref{fig:top_100k} and Fig.~\ref{fig:trigrams} show dataset coverage (y-axis) of top-n words and trigrams (x-axis) for each slimpajama category. Notably 10k trigrams, and 100k words consistently cover $>95\%$ of each slimpajama category.

\begin{table}[]
    \centering
    \begin{tabular}{c c}
    string & token id \\
    \hline\hline
`\_' & 49594 \\
\hline
`\_w'& 15997 \\
`W'& 40669 \\
`\_W'& 46854 \\
`w'& 63048 \\
\hline
`Wo'& 7411 \\
`\_wo'& 14297 \\
`\_WO'& 14883 \\
`wo'& 34034 \\
`\_Wo'& 39790 \\
`WO'& 44468 \\
\hline
`WOR'& 1916 \\
`\_WOR'& 6606 \\
`\_Wor'& 40813 \\
\hline
`\_Word'& 1971 \\
`Word'& 3212 \\
`\_word'& 14272 \\
`WORD'& 48022 \\
`word'& 49922 \\
\hline
`\_words'& 12555 \\
`words'& 28689 \\
`WORDS'& 32751 \\
`\_Words'& 37912 \\
`Words'& 51858\\
    \end{tabular}
    \caption{The 24 possible first tokens to construct uppercase and whitespace variations of ``\_words'', where ``\_'' denotes a whitespace. In total, there are 64 ways to write ``\_words'', which requires $32\cdot 6 + 32\cdot 5=342$ characters. 
    The tokenizer requires in total 194 tokens, of which 37 are unique, leading to an average (neglecting the occurrence frequencies) of $\approx3$ tokens per writing.}
    \label{tab:word_prefixes}
\end{table}

\section{More Benchmarks}
\label{app:benchmarks}
\begin{figure*}
    \centering
    \includegraphics[width=\linewidth,clip,trim=0 5cm 0 0]{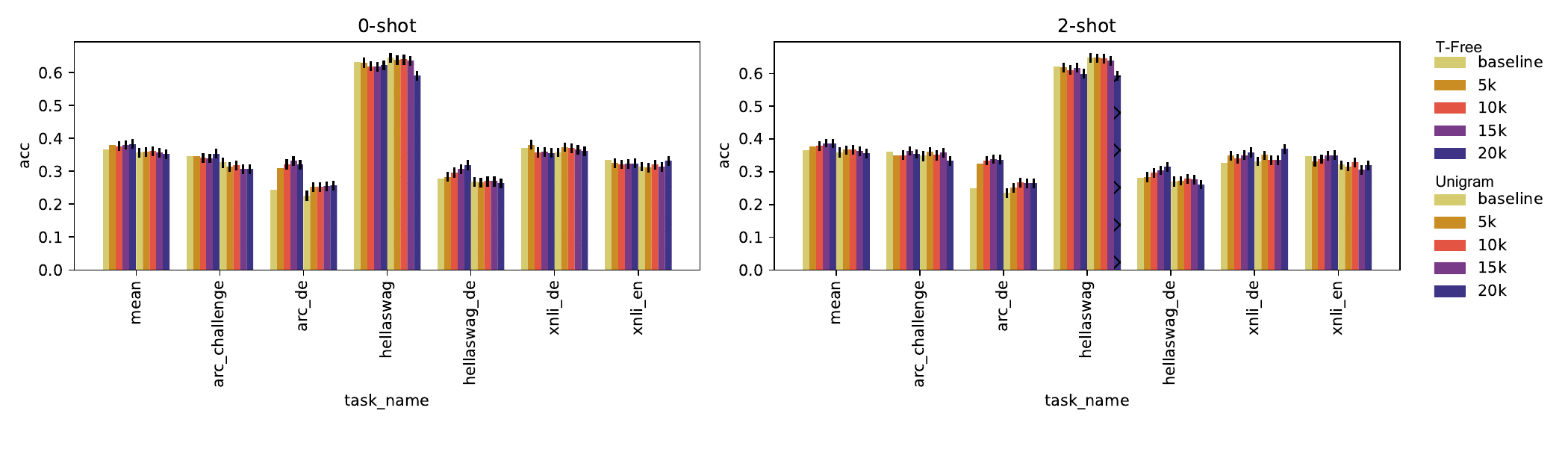}
    \includegraphics[width=\linewidth,clip,trim=0 5cm 0 0]{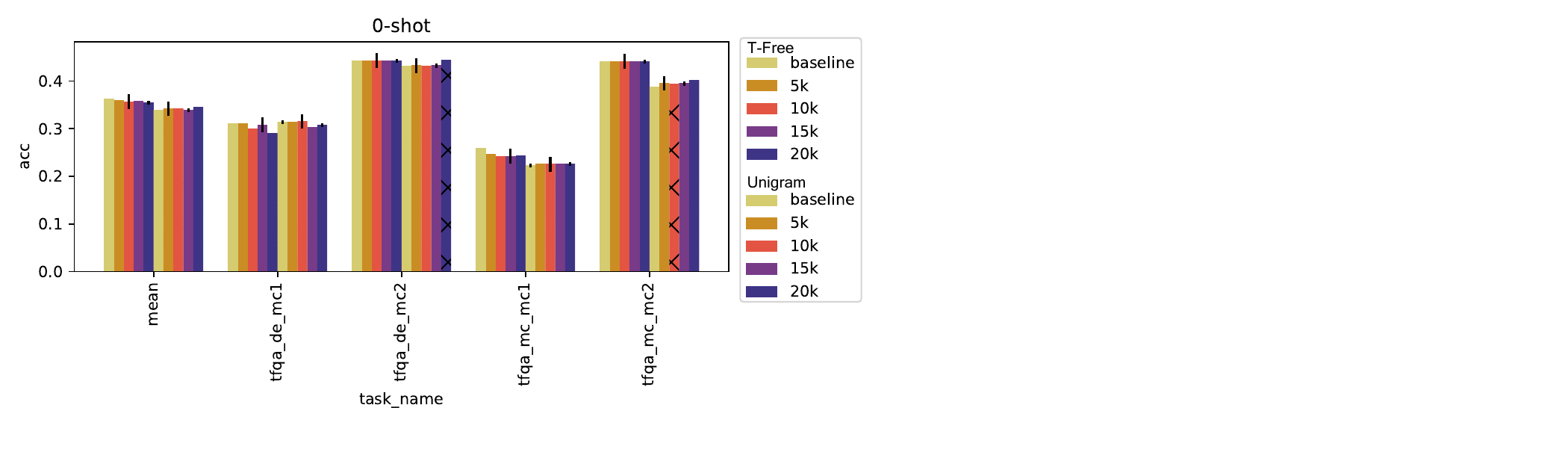}
    \includegraphics[width=\linewidth,clip,trim=0 5cm 0 0]{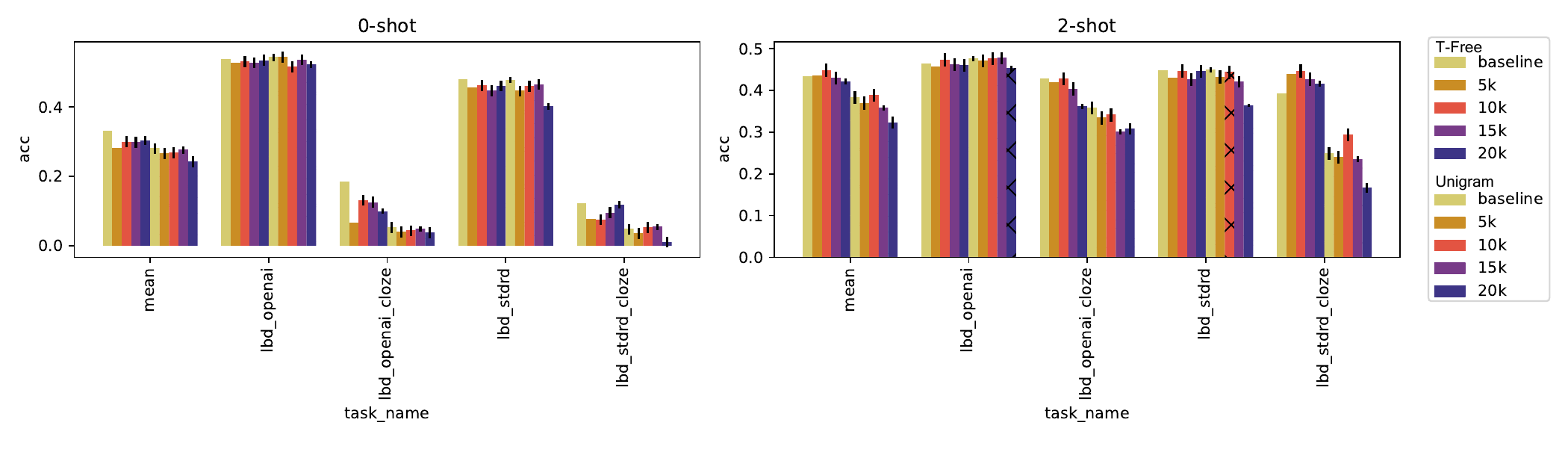}
    \includegraphics[width=\linewidth,clip,trim=0 5cm 0 0]{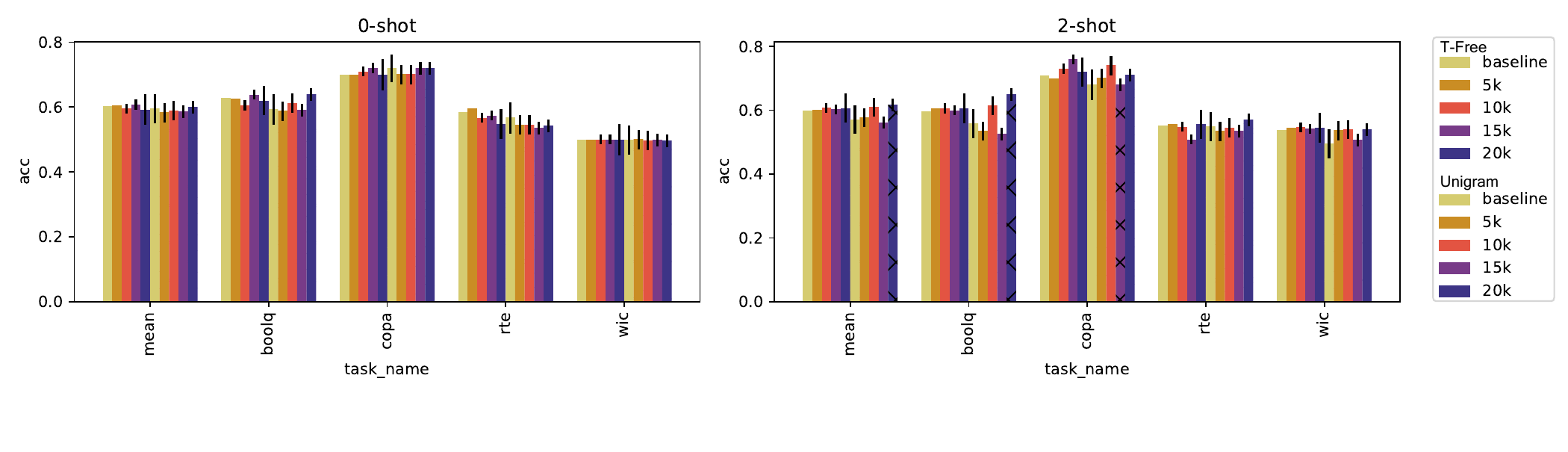}    
    \caption{Detailed benchmark results on evaluations of Sec.~\ref{sec:main_results}.}
    \label{fig:plot_all1}
\end{figure*}

\begin{figure*}
    \centering
    \includegraphics[width=\linewidth,clip,trim=0 5cm 0 0]{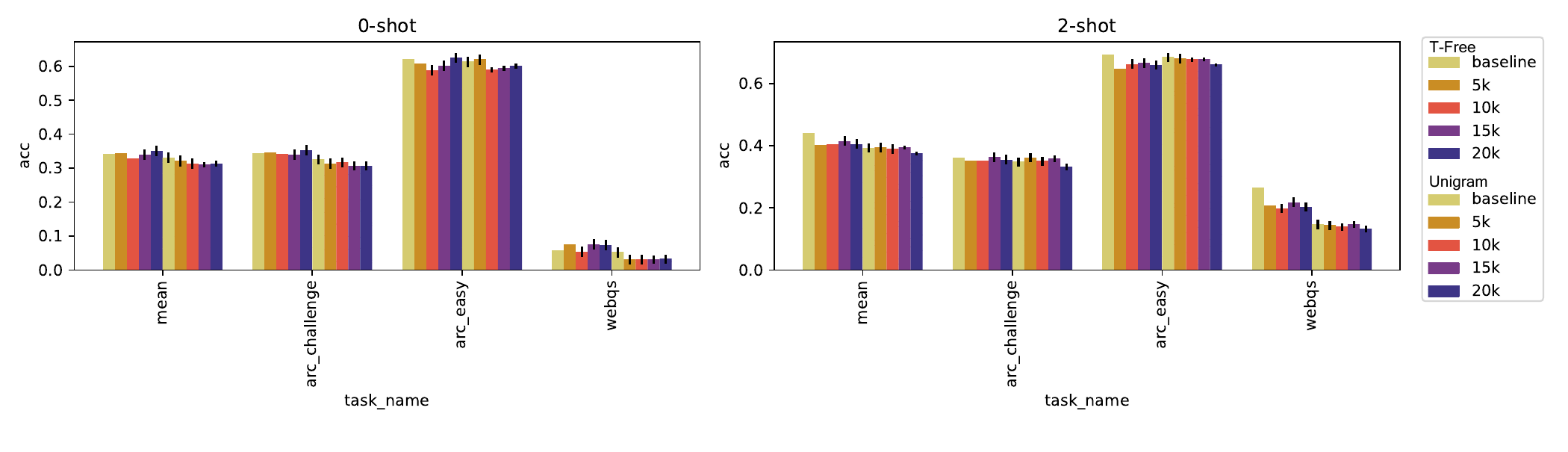}
    \includegraphics[width=\linewidth,clip,trim=0 5cm 0 0]{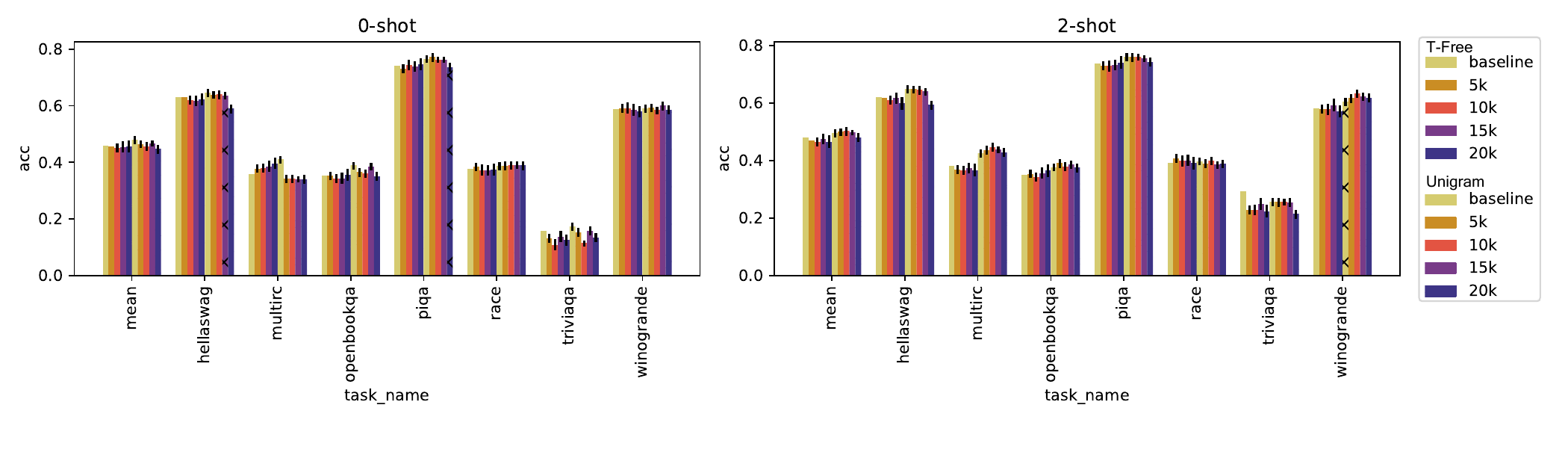}
    \caption{Further detailed benchmark results on evaluations of Sec.~\ref{sec:main_results}.}
    \label{fig:plot_all2}
\end{figure*}

We used the code of the eleuther eval harness, and evaluated each benchmark in 0-shot and 2-shot. All 18 benchmarks, namely
\textit{arc (easy and challenge), hellaswag, winogrande, triviaqa, xnli, truthfulqa, boolq, copa, openbook, piqa, multirc, lambada (openai and standard), race, rte, wic, webqs}
are visualized in Fig.~\ref{fig:plot_all1} and
Fig.~\ref{fig:plot_all2} for a baseline model trained on english slimpajama only and continued finetuning on german occiglot. 
Arc-challenge, hellaswag, xnli and truthfulqa are also evaluated in german translations.
Detailed numbers can be found in Tab.~\ref{tab:evalende},\ref{tab:evalen1} and \ref{tab:evalen2}.


\begin{table*}[]
\setlength{\tabcolsep}{5pt}
\small
    \centering
    \begin{tabular}{l l | r r r r r | r r r r r }
    \multicolumn{2}{c}{\multirow{2}{*}{\textbf{Model}}} & \multicolumn{5}{ c |}{\textbf{english benchmarks}} & \multicolumn{5}{ c }{\textbf{german benchmarks}}\\
        &  & $arc_{ch}$ & $hella$ & $xnli$ & $tf_{mc1}$ &  $tf_{mc2}$ &  $arc_{ch}$ & $hella$ & $xnli$ & $tf_{mc1}$ &  $tf_{mc2}$ \\ \hline \hline

        \parbox[t]{1mm}{\multirow{2}{*}{\rotatebox[origin=c]{90}{base}}} & 
  T-Free & 34.5/36.2 & 63.1/62.2 & 33.4/34.8 & 25.9/- & 44.1/- & 24.3/25.0 & 27.7/28.1 & 37.1/32.7 & 31.1/- & 44.3/-\\
 & Unigram & 32.6/34.8 & 64.5/64.8 & 31.4/32.1 & 22.3/- & 38.8/- & 22.5/23.5 & 26.8/27.0 & 35.7/33.1 & 31.3/- & 43.1/-\\
\hline\parbox[t]{1mm}{\multirow{2}{*}{\rotatebox[origin=c]{90}{5k}}} & T-Free & 34.7/35.1 & 63.0/61.9 & 32.5/33.2 & 24.7/- & 44.1/- & 30.9/32.5 & 28.3/28.4 & 38.1/34.9 & 31.1/- & 44.3/-\\
 & Unigram & 31.3/36.1 & 63.8/64.7 & 31.1/31.5 & 22.6/- & 39.6/- & 25.2/25.1 & 26.6/27.2 & 37.3/35.0 & 31.3/- & 43.2/-\\
\hline\parbox[t]{1mm}{\multirow{2}{*}{\rotatebox[origin=c]{90}{10k}}} & T-Free & 34.1/35.1 & 61.9/61.1 & 32.2/33.8 & 24.2/- & 44.1/- & 32.1/33.4 & 29.6/29.7 & 35.7/34.0 & 30.0/- & 44.3/-\\
 & Unigram & 31.7/35.0 & 63.9/64.5 & 32.0/32.9 & 22.5/- & 39.4/- & 25.1/26.6 & 26.9/27.8 & 37.0/33.5 & 31.5/- & 43.1/-\\
\hline\parbox[t]{1mm}{\multirow{2}{*}{\rotatebox[origin=c]{90}{15k}}} & T-Free & 33.9/36.3 & 61.8/61.8 & 32.3/34.9 & 24.2/- & 44.1/- & 33.2/33.9 & 30.7/30.4 & 35.9/35.1 & 30.8/- & 44.3/-\\
 & Unigram & 30.6/35.8 & 63.5/63.9 & 31.4/30.5 & 22.6/- & 39.4/- & 25.3/26.4 & 26.9/27.6 & 36.5/33.4 & 30.2/- & 43.2/-\\
\hline\parbox[t]{1mm}{\multirow{2}{*}{\rotatebox[origin=c]{90}{20k}}} & T-Free & 35.3/35.5 & 62.2/59.9 & 32.4/35.1 & 24.4/- & 44.1/- & 32.0/33.7 & 31.9/31.5 & 35.6/35.9 & 29.1/- & 44.2/-\\
 & Unigram & 30.6/33.2 & 58.9/59.3 & 33.1/31.9 & 22.6/- & 40.2/- & 25.7/26.5 & 26.3/26.1 & 36.0/36.9 & 30.7/- & 44.4/-

    \end{tabular}
    \caption{Accuracy scores of english and german translated benchmarks for continued pre-training. First value denotes 0-shot, second value 2-shot (if available). Notably, the T-Free baseline model slightly outperforms (or performs on par with) the Unigram baseline model on all of these tasks. On german evals of arc and hellaswag, the T-Free baseline  outperforms Unigram, and achieves larger gains during continued training on the german/ english data mix. The german versions of xnli and truthfulqa mostly remain unchainged.}
    \label{tab:evalende}
    \vskip 1.em
\end{table*}

\vspace{-10pt}\begin{table*}[]
\small
    \centering
    \begin{tabular}{l l | r r r r r r r r}
        \multicolumn{2}{c}{\multirow{2}{*}{\textbf{Model}}} & \multicolumn{8}{ c }{\textbf{english benchmarks}} \\
    & & 
        $arc_{ez}$ &  $boolq$ & $copa$ & $wino$ & $obook$ &  $piqa$ &  $trivia$ &  $mrc$   \\ \hline \hline
    \parbox[t]{1mm}{\multirow{2}{*}{\rotatebox[origin=c]{90}{base}}} & T-Free & 62.1/69.4 & 62.9/59.8 & 70.0/71.0 & 58.7/58.3 & 35.4/35.0 & 74.0/73.8 & 15.7/29.3 & 36.0/38.1\\
 & Unigram & 61.3/68.4 & 59.2/55.8 & 72.0/68.0 & 59.0/60.4 & 38.8/37.6 & 76.5/76.0 & 17.2/25.6 & 40.9/42.4\\
 \hline\parbox[t]{1mm}{\multirow{2}{*}{\rotatebox[origin=c]{90}{5k}}}& T-Free & 60.8/64.8 & 62.6/60.7 & 70.0/70.0 & 59.0/57.9 & 35.2/35.4 & 73.0/73.0 & 13.1/22.9 & 37.8/36.9\\
 & Unigram & 62.0/68.0 & 58.7/53.5 & 70.0/70.0 & 59.2/61.6 & 36.4/39.0 & 77.1/75.9 & 15.2/25.5 & 34.1/43.5\\
 \hline\parbox[t]{1mm}{\multirow{2}{*}{\rotatebox[origin=c]{90}{10k}}}& T-Free & 58.8/66.3 & 60.5/60.6 & 71.0/73.0 & 59.2/57.8 & 34.2/34.2 & 74.4/72.9 & 10.9/22.8 & 38.0/36.6\\
 & Unigram & 59.0/67.7 & 61.2/61.4 & 70.0/74.0 & 58.4/63.2 & 36.0/37.8 & 76.2/75.9 & 11.4/25.6 & 34.1/44.6\\
\hline\parbox[t]{1mm}{\multirow{2}{*}{\rotatebox[origin=c]{90}{15k}}} & T-Free & 60.2/66.6 & 63.8/60.0 & 72.0/76.0 & 58.6/59.3 & 34.4/35.6 & 73.9/73.3 & 13.8/24.9 & 38.6/37.3\\
 & Unigram & 59.4/67.8 & 59.0/52.6 & 72.0/68.0 & 59.9/62.2 & 38.4/38.6 & 76.2/75.5 & 15.7/25.4 & 34.0/43.8\\
\hline\parbox[t]{1mm}{\multirow{2}{*}{\rotatebox[origin=c]{90}{20k}}} & T-Free & 62.5/66.0 & 62.0/62.6 & 70.0/72.0 & 57.9/57.1 & 35.6/36.5 & 74.6/74.1 & 12.5/22.4 & 39.7/36.7\\
 & Unigram & 60.1/66.0 & 63.8/62.9 & 72.0/71.0 & 58.5/61.8 & 35.0/37.4 & 73.5/74.2 & 13.4/21.3 & 34.0/42.7

    \end{tabular}
    \caption{Accuracy scores of english benchmarks for continued pre-training. First value denotes 0-shot, second value 2-shot. Notably, the T-Free model performs on par to the Unigram  model on all of these tasks, throughout the entire continued training. }
    \label{tab:evalen1}
    \vskip 1.em
\end{table*}

\begin{table*}[]
\small
    \centering
    \begin{tabular}{l l | r r r r r r r r}
        \multicolumn{2}{c}{\multirow{2}{*}{\textbf{Model}}} & \multicolumn{8}{ c }{\textbf{english benchmarks}} \\
    & &  $lbd^{oai}$ & $lbd^{oai}_{clz}$  & $lbd^{stdr}$  &  $lbd^{stdr}_{clz}$ &  $race$ &  $rte$ &  $wic$ &  $webqs$ \\ \hline \hline
     \hline\parbox[t]{1mm}{\multirow{2}{*}{\rotatebox[origin=c]{90}{base}}} & T-Free & 53.9/46.4 & 18.5/42.9 & 48.0/44.8 & 12.2/39.3 & 37.8/39.1 & 58.5/55.2 & 50.0/53.8 & 5.7/26.5\\
 & Unigram & 54.3/47.6 & 5.2/35.8 & 47.8/44.9 & 4.7/24.9 & 38.6/39.7 & 56.7/54.9 & 49.8/49.5 & 5.2/14.6\\
 \hline\parbox[t]{1mm}{\multirow{2}{*}{\rotatebox[origin=c]{90}{5k}}}& T-Free & 52.7/45.7 & 6.6/41.9 & 45.6/43.0 & 7.8/44.0 & 38.4/40.8 & 59.6/55.6 & 50.0/54.5 & 7.6/20.8\\
 & Unigram & 54.4/47.1 & 3.9/33.4 & 44.6/43.2 & 3.5/24.0 & 38.7/38.9 & 54.5/53.4 & 50.0/53.6 & 3.0/14.3\\
 \hline\parbox[t]{1mm}{\multirow{2}{*}{\rotatebox[origin=c]{90}{10k}}}& T-Free & 53.1/47.3 & 13.1/42.8 & 46.2/44.7 & 7.5/44.7 & 37.3/39.9 & 56.7/54.9 & 50.0/54.7 & 5.4/19.8\\
 & Unigram & 51.6/47.6 & 4.3/34.1 & 46.0/44.3 & 5.2/29.4 & 39.0/39.9 & 54.5/54.5 & 49.7/53.9 & 3.0/13.8\\
 \hline\parbox[t]{1mm}{\multirow{2}{*}{\rotatebox[origin=c]{90}{15k}}}& T-Free & 52.8/46.2 & 12.5/40.4 & 44.7/42.6 & 9.5/42.7 & 37.1/40.0 & 57.4/50.9 & 50.0/54.2 & 7.6/21.8\\
 & Unigram & 53.6/47.7 & 4.8/30.1 & 46.5/42.0 & 5.4/23.5 & 39.0/38.5 & 53.4/53.4 & 49.8/50.6 & 3.0/14.6\\
\hline\parbox[t]{1mm}{\multirow{2}{*}{\rotatebox[origin=c]{90}{20k}}} & T-Free & 53.5/46.0 & 9.9/36.2 & 46.1/44.6 & 11.9/41.6 & 37.4/39.1 & 54.9/55.6 & 50.0/54.5 & 7.4/20.3\\
 & Unigram & 52.2/45.2 & 3.7/30.8 & 40.1/36.4 & 1.0/16.6 & 38.8/38.8 & 54.2/57.0 & 49.5/53.9 & 3.2/13.1
    \end{tabular}
    \caption{Accuracy scores of english benchmarks for continued pre-training. First value denotes 0-shot, second value 2-shot. The T-Free model performs on par with the Unigram model on all of these tasks, throughout the entire continued training. Notably, the clozed variants of lambada are most fragile, at which T-Free outperforms.}
    \label{tab:evalen2}
    \vskip 1.em
\end{table*}


\end{document}